\documentclass[letterpaper, 10 pt, conference]{ieeeconf} 

\usepackage{CJKutf8} 
\usepackage{amsmath}
\usepackage{amsfonts}
\usepackage{graphicx}
\usepackage{subfigure}
\usepackage{hyperref}
\newcommand{\RefSec}[1]{Section~\ref{#1}}
\newcommand{\RefEq}[1]{Equation (\ref{#1})}

\newcommand{\RefFig}[1]{Fig. \ref{#1}}

\IEEEoverridecommandlockouts                              

\usepackage{setspace}
\renewcommand{\baselinestretch}{0.98} 

\title{\LARGE \bf
Locomotion Generation for a Rat Robot based on Environmental Changes via Reinforcement Learning
}

\author{Xinhui Shan$^{1,\dagger}$, Yuhong Huang$^{1,\dagger}$, 	Zhenshan Bing$^{1}$, Zitao Zhang$^{2}$, Xiangtong Yao$^{1}$, Kai Huang$^{2}$, Alois Knoll$^{1}$
\thanks{$^{1}$Authors from the Technical University of Munich, Munich, Germany {\tt\small yuhong.huang@tum.de}}%
\thanks{$^{2}$Authors from the Sun Yat-Sen University, Guangdong, China}
 \thanks{$\dagger$ These authors contributed equally to this work}%
}

\begin{document}

\maketitle
\thispagestyle{empty}
\pagestyle{empty}

\begin{abstract}
This research focuses on developing reinforcement learning approaches for the locomotion generation of small-size quadruped robots. The rat robot NeRmo is employed as the experimental platform.  Due to the constrained volume, small-size quadruped robots typically possess fewer and weaker sensors, resulting in difficulty in accurately perceiving and responding to environmental changes. In this context, insufficient and imprecise feedback data from sensors makes it difficult to generate adaptive locomotion based on reinforcement learning. To overcome these challenges, this paper proposes a novel reinforcement learning approach that focuses on extracting effective perceptual information to enhance the environmental adaptability of small-size quadruped robots. According to the frequency of a robot's gait stride, key information of sensor data is analyzed utilizing sinusoidal functions derived from Fourier transform results. Additionally, a multifunctional reward mechanism is proposed to generate adaptive locomotion in different tasks. Extensive simulations are conducted to assess the effectiveness of the proposed reinforcement learning approach in generating rat robot locomotion in various environments. The experiment results illustrate the capability of the proposed approach to maintain stable locomotion of a rat robot across different terrains, including ramps, stairs, and spiral stairs. The experiment video is shown at \underline{\url{https://youtu.be/NYLknYGDYxc}}.
\end{abstract}

\section{Introduction}\label{intro}
Quadruped robots possess remarkable freedom of movement in three-dimensional space, enabling them to adapt to various environmental changes \cite{gong2010review, biswal2021development}. Against different environments, researchers have to develop specialized model-based controllers to generate robot motions. However, the uncertainty of the real world causes significant complexity in designing the controllers \cite{bellegarda2022robust}. To mitigate this issue, model-free Reinforcement Learning (RL) approaches emerge as a potential approach for controlling robots in diverse environments \cite{degris2012model}.

\begin{figure}[!t]
\centering
\includegraphics[width=.9\columnwidth,trim=3cm 17.5cm 0 3.6cm ,clip]{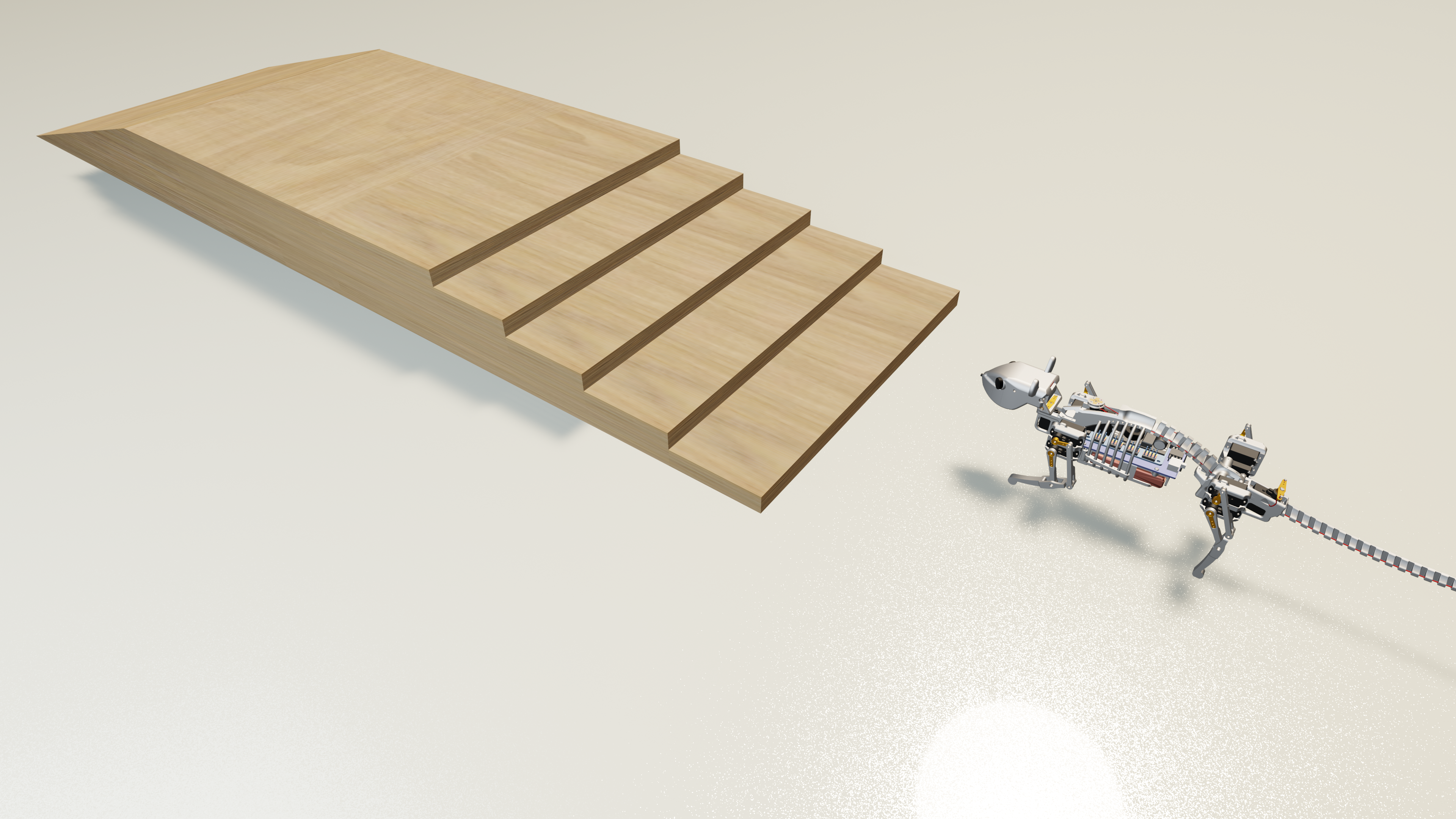}
\caption{A rat-like robot attempts to ascend the stairs.}
\label{fig:case}
\vspace{-1.0em} 
\end{figure}

Learning from the interaction between robots and their environments, RL approaches are able to autonomously generate adaptability quadrupedal locomotion based on real-time feedback information \cite{luo2020carl, shi2022reinforcement, aractingi2023controlling}. For instance, to enhance the generality and robustness of quadrupedal locomotion, the ANYmal robot acquired its gait without specific environmental data through RL \cite{hwangbo2019learning, lee2020learning}. Furthermore, to address the diverse requirements of complex tasks, hierarchical RL has been utilized to decompose tasks and guide robot locomotion \cite{lee2020learning, jain2019hierarchical, jenelten2024dtc}. Many prior research endeavors have simplified the design of robot control models and bolstered adaptability to unknown environments through RL. However, such approaches often rely on the variability in data obtained from various sensors of full-size quadruped robots. For small-size quadruped robots, volumetric constraints limit sensor configurations, thereby posing difficulties in understanding environmental changes (as illustrated in \RefFig{fig:case}).  In this case, RL approaches encounter considerable difficulty in generating adaptive locomotion for small-size quadruped robots in diverse environments.

To address various environmental changes, small-size quadruped robots employ additional Degrees Of Freedom (DOF) to improve their locomotive capabilities \cite{shi2020implementing, bing2023lateral}. For instance, the rat robot ``NeRmo" integrates a soft actuated spine, enabling it to dynamically adapt to varying terrains \cite{lucas2019development}. Through the coordinated mechanism of the spinal flexion alongside limb locomotion, ``NeRmo" has demonstrated enhanced flexibility, achieving reduced turning radius and accelerated speeds \cite{bing2023lateral, huang2022enhanced}. Similarly, ``SQuRo-S'' generates diverse special behaviors by coupling compliant spinal bending that can significantly improve its environmental adaptability \cite{wang2024bioinspired}. Although additional DOFs in the spine enhance greater flexibility in robot locomotion, they cause heightened complexity in the design of control models. Moreover, in the absence of feedback information,  the applications of small-size quadruped robots are limited to specific tasks with certain environmental changes.

\begin{figure*}[ht]
    \centering
    \includegraphics[width=.9\linewidth, trim=1cm 3cm 4cm 3cm, clip]{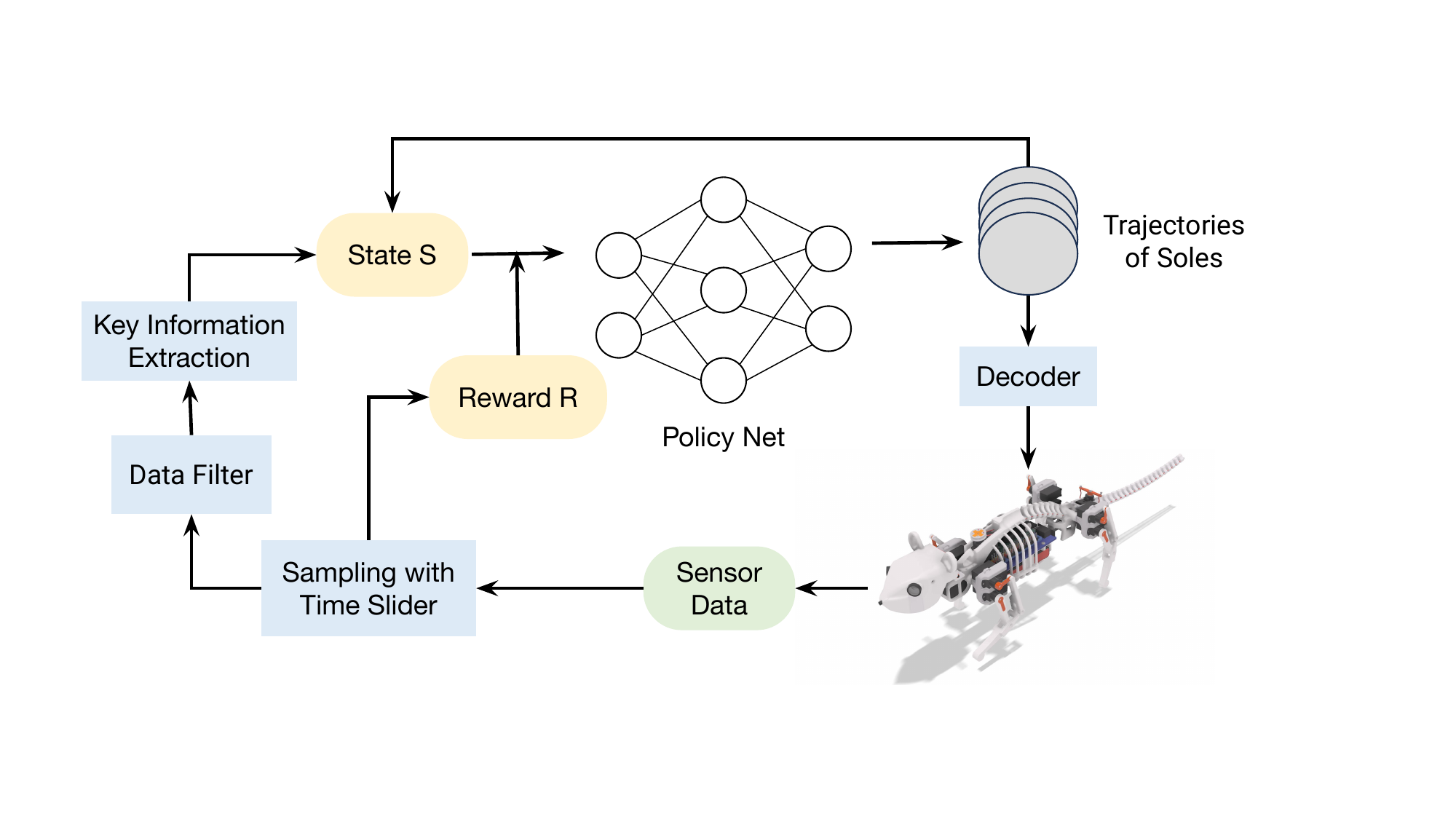}
    \caption{Architecture for applying RL into control of a rat robot. The blue boxes represent the data operation. The sensor data in green boxes is directly observed during robot work. The gray box is the action generated by RL's policy net. The yellow boxes are the necessary input of the policy net.}
    \vspace{-0.5em} 
    \label{fig:archi}
\end{figure*}
Despite the effectiveness of RL in enhancing environmental adaptability, small-size quadruped robots focus on model-based approaches to generate locomotion. The reasons are multi-fold. Firstly, the movement instability in small-size quadruped robots often leads to considerable sensor noise. Secondly, the limited physical volume of these robots poses challenges in accommodating diverse sensors necessary for capturing detailed environmental changes. Thirdly, the scarcity and weakness of feedback information intensify the complexities of designing effective reward mechanisms for the robot's actions in response to environmental changes.

To handle these challenges, this paper proposes a novel reinforcement learning approach that focuses on extracting effective perceptual information to enhance the environmental adaptability of small-size quadruped robots. Leveraging our prior work on the rat robot ``NeRmo'' \cite{bing2023lateral}, we employ it as the experimental platform in this study.

Our main contributions are summarized as follows:
\begin{itemize}
     \item To describe environmental changes using data from weak sensors, this study focuses on extracting essential information based on the frequency of a gait stride that can reduce noise during walking.
     \item To illustrate differences among environmental changes, this research integrates feedback information by combining sinusoidal functions derived from Fourier transform analysis. This integration aims to extract key information to construct the state space utilized in reinforcement learning (RL).
     \item To address the diverse environmental changes, this paper designs a multifunctional reward mechanism that can be easily tailored to different task requirements.
\end{itemize}

\section{Overview}\label{overview}

\RefFig{fig:archi} illustrates the architecture of the proposed RL approach. Initially, we analyze the data features during robot locomotion. Given the characteristic of significant sensor noise during the locomotion of small-size robots, data sampling with time sliders infers effective information based on the continuity of robot motion. Subsequently, based on the sampled information, reward signals are defined to depict the effectiveness of the robot's current actions in the environment. 

The robot then generates real-time robot states involving the task based on the sampled information. To mitigate the impact of noise on the description of robot states, the sampled information is filtered based on the robot's gait stride frequency. To enrich the representation of the environment with sensor information, this data is integrated through sinusoidal functions derived from Fourier transform results. Parameters of sinusoidal functions characterize differences in the robot's states in different environments. The real-time robot states also need to cover historical action information associated with the robot's locomotion.

Based on the robot's states and action rewards, the robot utilizes a policy network for learning and generates the next actions. These actions are defined based on the trajectory of limb soles and translated into specific control parameters for each servo motor through a decoder to drive the robot's motion.

Finally, the robot will interact with the environment using newly generated actions and observe new sensor information. Through this iterative process, the robot continuously explores and discovers optimized actions to adapt to different environments through trial and error.

\section{Environment Perception}\label{env}


To address the perception limitations in small-sized robots, this section outlines a sensor data analysis strategy employing the Fast Fourier Transform (FFT) and uses a combination of sinusoidal functions to describe environmental variations encountered in various tasks.

\begin{figure*}[ht]
    \centering
    \includegraphics[width=.95\linewidth, trim=0cm 0.2cm 0cm 0cm, clip]{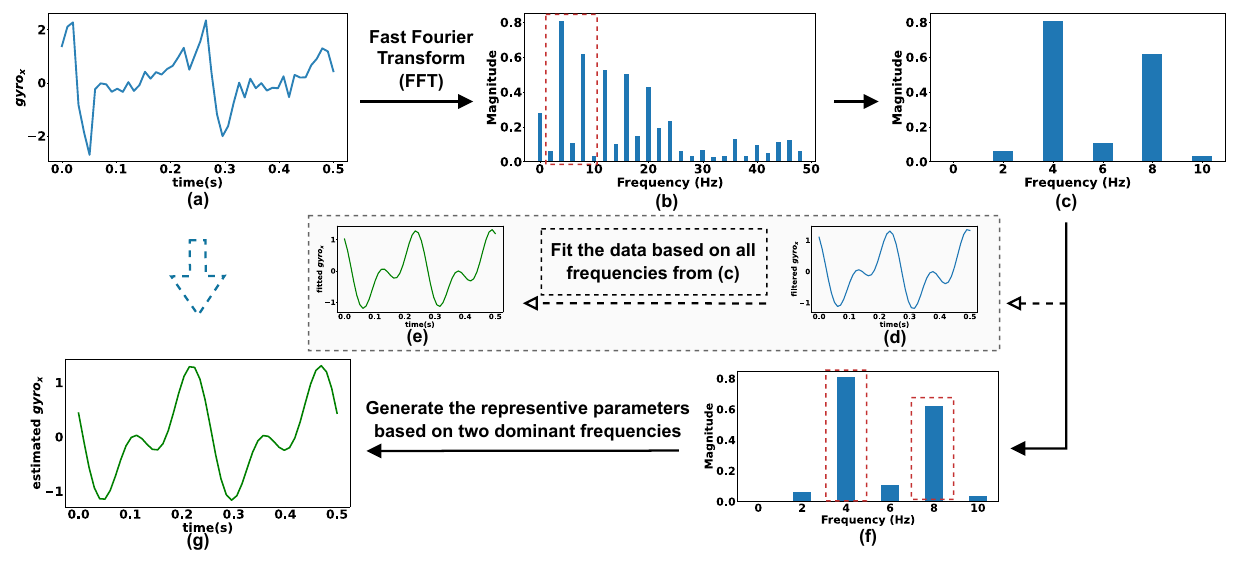}
    \caption{Data processing for analyzing sensor data. This figure outlines analyzing gyroscopic data along the x-axis in one period T, sampled at a 100Hz sampling rate. (a) displays the raw data. In (b), we apply the Fast Fourier Transform (FFT) to identify frequency components. (c) filters out frequencies outside the 0.1Hz-10Hz range to weaken noise. (d) shows the filtered time signals transformed by the Inverse Fast Fourier Transform (IFFT). In (e), we fit all filtered time signals referring to the \RefEq{sum_fuc}. (g) shows the data fitted using only the significant frequencies from (f) according to \RefEq{two_sin}, which can also effectively represent environmental changes.}
    \vspace{-0.8em}
    \label{fig:fft}
\end{figure*}

\subsection{Sensor Key Information Extraction}\label{t_sensor}


Given the compact size of the rat robot NeRmo, embedding a wide range of sensors is difficult, resulting in limited perception. Besides, the locomotion of NeRmo easily introduces numerous noise within the sensor data, further complicating the analysis process. In response to these challenges, our approach focuses on analyzing environmental changes based on extracted key information from the constrained sensor data.

Limited by the volume, the effective sensor of NeRmo is only the small Inertial Measurement Unit (IMU) IMU with type ``GY-521'', consisting of three-dimensional acceleration vectors $\overrightarrow{acc}$ and angular velocity vectors $\overrightarrow{gyro}$. With the fixed time sliders, the observation space $O_{t'}$ from the interaction between robot and environment is denoted as
\begin{equation}
\begin{aligned}
\overrightarrow{acc} &= [\ acc_x,\ acc_y,\ acc_z\ ], \\
\overrightarrow{gyro} &= [\ gyro_x,\ gyro_y,\ gyro_z\ ], \\
o_t &= [\ \overrightarrow{acc},\ \overrightarrow{gyro}\ ], \\
O_{n} &= \{ o_t, t \in [nT, (n+1)T], n \in \mathbb{N} \},
\end{aligned}\label{data_set}
\end{equation}
where $T$ denotes the period of a gait stride. The effective sensory information extraction is shown in \RefFig{fig:fft}(a) to (d), taking the gyroscopic data along the x-axis as an example.

\RefFig{fig:fft}(a) depicts the original sensor data obtained from NeRmo, interfered with by frequent shakes during robot movement. As the robot maintains stable locomotion, its perceptual information theoretically exhibits periodicity similar to its gait strides. Building upon this insight, we employ the Fast Fourier Transform (FFT) to convert the time-domain sensor data into frequency-domain representation, benefiting analysis of the information perceived by the robot in response to environmental changes, as depicted in \RefFig{fig:fft}(b).
The process of data transformation using FFT is described by
\begin{equation}\label{fft_eq}
 F(f) = \int_{nT}^{(n+1)T} o(t)e^{-2\pi i ft} dt,  f\in[0.1, 10],
\end{equation}
where $F(f)$ denotes the complex amplitude of frequency $f$, and $i$ represents the imaginary unit. In the context of NeRmo's applications, considering the physical constraints inherent in the robot, the frequency of gait stride typically falls within the range of [0.5 Hz, 2 Hz]. Accordingly, the frequency of the effectively sensed data should be around the frequency of a gait stride. Therefore, to enhance the robustness of robotic perception under these conditions, the frequency of sensed data is strategically selected within the range of [0.1 Hz, 10 Hz], as depicted in \RefFig{fig:fft}(c).

\begin{figure*}[ht]
    \centering
    \includegraphics[width=.95\linewidth]{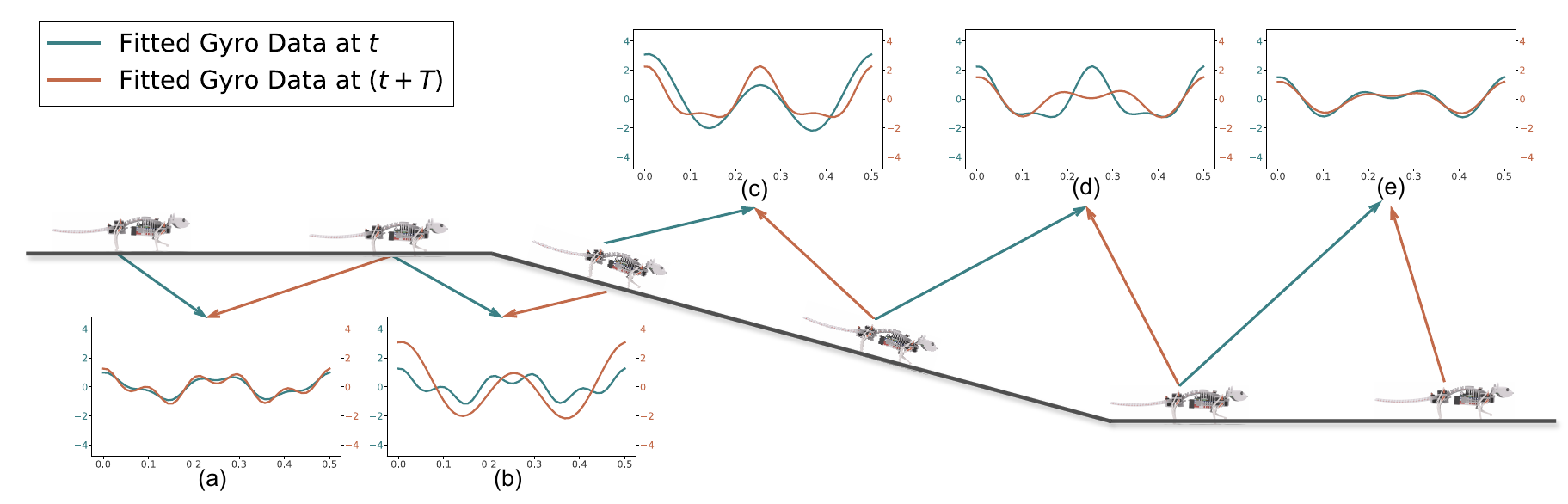}
    \caption{Gyroscope data variation on the x-axis of NeRmo. During NeRmo's ramp descent, six time slots are observed: (a) Data on flat terrain, (b) Data variation at the start of the descent, (c) Data during the descent, (d) Data variation transitioning from descent to flat terrain, (e) Data on the lower flat terrain.}
    \vspace{-0.5em} 
    \label{fig:5trans}
\end{figure*}

Building upon this, the sensed data with noise filtered out is shown in the \RefFig{fig:fft}(d). To transition from frequency-domain data back to the time domain and generate filtered real-time perceptual information, we employ the Inverse Fourier Transform (IFFT).  Considering \RefEq{fft_eq}, the transformation process is 
\begin{equation}\label{ifft}
o'(t) = \int F(f)e^{2\pi i ft}df,
\end{equation}
where $o'(t)$ represents the sensed data with noise filtered out. This method aims to mitigate the influence of noise on perceptual information, which is caused by the unstable movement characteristic of small-sized quadruped robots.


To describe the environmental changes observed by the robot utilizing filtered sensed data, we employ the Fourier series to characterize the perception information. This approach enables us to integrate the perception information within a set of predetermined functions, effectively representing environmental variations. Drawing upon the Fourier series, the perception information produced by the IFFT can be regarded as a composite of numerous sinusoidal functions, each corresponding to its inherent periodicity. In other words,
\begin{equation}\label{sum_fuc}
o'(t)=\sum_{j}^{m} F({f}_{j})\sin{(2 \pi f_{j}t+\phi_{j})}+C_{j}, \quad m \in \mathbb{N}.
\end{equation}
In \RefEq{sum_fuc}, the terms $f_j$ and $F(f_j)$ are defined and computed in \RefEq{fft_eq}. The variable $\phi_j$ signifies the phase shifts, while the constant $C_j$ signifies the offset. The parameter $m$ signifies the number of distinct sinusoidal signals utilized to compose $o'(t)$.
When the parameter $m$ in \RefEq{sum_fuc} reaches the maximum number of frequencies based on discrete sampling,  the result of $o'(t)$ (shown \RefFig{fig:fft}(e)) is almost the same with \RefFig{fig:fft}(d). While this approach can effectively generate high-dimensional signals to accurately express environmental changes, it also leads to heavy computational overhead. It must be admitted that significant time costs from the computation are impractical for the control of real-time systems such as robots. As shown in \RefFig{fig:fft}(f), We observe that there are always two signals that play leading roles in the integration of all signals. Thereby, to mitigate computational challenges and enhance data efficiency, we designed a pair of sinusoidal functions that combine these two main signals. Consequently, \RefEq{sum_fuc} can be simplified as
\begin{equation}\label{two_sin}
\begin{aligned}
o'(t) &=  F(f_1) \sin( 2 \pi f_1 t + \phi_1) \\
& + F(f_2) \sin(2 \pi f_2 t + \phi_2) + C', \\ 
&\text{for} \ \forall f, \ F({f}_1)>F({f}_2)>F({f}_j).
\end{aligned}
\end{equation}
Where $f_1$ and $f_2$ represent the frequencies of the two signals with the maximum amplitudes. The constant $C'$ is formed by the addition of $C_1$ and $C_2$ which are related to $f_1$ and $f_2$ respectively. \RefFig{fig:fft}(g) showcases the functional graph of \RefEq{two_sin}. Compared to \RefFig{fig:fft}(e), although \RefFig{fig:fft}(g) exhibits some numerical deviations, their patterns of variation show a high degree of consistency. This consistency is precisely what we need, as in the process of the robot responding to environmental changes, we are more concerned with the changes in perception information. According to this, we will be able to utilize specific mathematical models to model the perception information obtained by sensors within a time slider. This innovative approach significantly captures the environmental changes, benefiting RL's approach to analyzing the interaction between the robot and the environment.



\subsection{Discussion for Describing Environmental Changes}
The periodicity of robot locomotion results in stable rhythmic signals in NeRmo's perception information while walking on flat terrain. Slight differences in these signals emerge when encountering varied terrains. Detecting such slight differences rapidly without using the Fourier transform can be challenging due to the instability of time-domain signals. Through basic signal processing techniques, these slight changes may be erroneously classified as noise. Therefore, utilizing the approach introduced in \RefSec{t_sensor} to describe the modification of the robot's environment in real-time.

Utilizing \RefEq{two_sin}, the differences observed as the robot traversed uneven terrains are depicted in \RefFig{fig:5trans}. In \RefFig{fig:5trans}(a), the rat robot maintains stable locomotion on flat terrain, exhibiting minimal deviations. However, as the robot descends the ramp in \RefFig{fig:5trans}(b), notable signal alterations in response to the changing environment. Subsequently, as the rat robot continues walking down during a consistent environment, its perceptional information becomes stable, as depicted in \RefFig{fig:5trans}(c). Upon completing the ramp descent, the rat robot resumes ground locomotion, reflected in the significant differences in \RefFig{fig:5trans}(d). Eventually, the signals on flat terrain gained stability, as shown in \RefFig{fig:5trans}(e).

In summary, the trend variations observed in the fitted data closely correspond to the interaction between the robot and its environment. In this context, the ability of the proposed approach to effectively capture and represent environmental changes is validated.

\section{Reinforcement Learning Application in NeRmo}\label{rl}

This section focuses on the design of action space and reward mechanisms for RL. We describe the action space based on the trajectory of limb soles using polar coordinates. Additionally, we develop multifunctional reward mechanisms aimed at enhancing the robot's adaptability to a series of environmental and task conditions.

\subsection{Action Space}


To adapt to diverse environments, the difficulty for the robot lies in how to generate adaptable gaits based on environmental changes. The Inverse Kinematics (IK) of a robot is typically well-defined during its design phase. With a specified IK, the robot's gait can be represented as the trajectory of its limb soles. In essence, to effectively adapt to various environments, the robot must dynamically generate suitable trajectories for each limb sole in real time. Consequently, the action space in our (RL) is defined by these trajectories of limb soles.

Building upon our previous research \cite{bing2023lateral}, we present the trajectory of each limb sole precisely using the polar coordinate system. Considering the periodicity and physical constraints of each limb motion, an action within the action space can be defined as:
\begin{equation}\label{action}
\overrightarrow{a} = [\overrightarrow{\rho},\overrightarrow{\theta}, \overrightarrow{f'}],
\end{equation}
where $\rho$ and $\theta$ refer to the trajectory of the limb sole and $f'$ encompasses the frequency of a gait stride. To generate quadrupedal locomotion, $\overrightarrow{\rho}$, $\overrightarrow{\theta}$, and $\overrightarrow{f'}$ are four-dimensional vectors, each reflecting the control parameters for the four limbs.

\subsection{State Space}

As discussed in \RefSec{env}, the observed environmental changes within a time slider can be represented as a composite of two sinusoidal functions. This implies that the environmental variations depicted by one of the dimensions of data from the sensor can be delineated by the parameters of such two functions. Coupling this insight with \RefEq{two_sin}, the environmental changes are
\begin{equation}\label{obs1}
\begin{aligned}
Env(x)= &[F({f}_{1}(x)), \ F({f}_{2}(x)), \\
&{f}_{1}(x), \ {f}_{2}(x), \ \phi_1(x), \ \phi_2(x), \ C(x)],
\end{aligned}
\end{equation}
where $x$  denotes a specific dimension of the sensed data as outlined in \RefEq{data_set}. To comprehensively characterize environmental changes, we incorporate data from the accelerometer along the y-axis and all gyroscope measurements, where the y-axis aligns with the robot's forward direction. Taking into account the historical actions, the state of the robot is defined as:
\begin{equation}\label{obs}
\begin{aligned}
\mathcal{S} = &[Env(gyro_x), \  Env(gyro_y), \\
&Env(gyro_z), \ Env(acc_y), \ \overrightarrow{a}].
\end{aligned}
\end{equation}
For the state $S$ of $n^{th}$ time slider, $Env(gyro_x)$, $Env(gyro_y)$, $Env(gyro_z)$, and $Env(acc_y)$ are generated by the sensed data in $n^{th}$ time slider, while $\overrightarrow{a}$ is the executed action in $(n-1)^{th}$  time slider. These states across all time sliders within the training iteration collectively form the state space for RL.

\subsection{Multifunctional Reward Function}


To adapt to diverse environments, the robot needs to optimize its actions based on different environmental changes. Thus, a multifunctional reward mechanism for RL becomes essential to optimize robot actions for environmental adaptivity. This reward mechanism will integrate multiple factors of robotic behavior, thereby enhancing its learning process. In this paper, the overall reward $R$ is expressed as:
\begin{equation}\label{re_all}
R = 
\begin{cases} 
P_f&, \text{if }  |\omega_d| > \omega_f \\
r(s_x) + r(s_y) + r(s_z) + \sum L_i&, else
\end{cases}
\end{equation}
The parameter $\omega_d$ represents the gyroscope data along the directional axis. If $\omega_d$ exceeds a predefined threshold $\omega_f$, it indicates a risk of the robot rolling over. In this case, the action taken makes continuous walking unfeasible, resulting in a substantial penalty $P_f$ (assigned as -10). Consequently, the robot halts its current learning iteration and undergoes a reset process.

\begin{figure}[t]
    \centering
    \subfigure[Reward function of ramps and stairs hybrid environment.]{
		\includegraphics[width=.45\columnwidth]{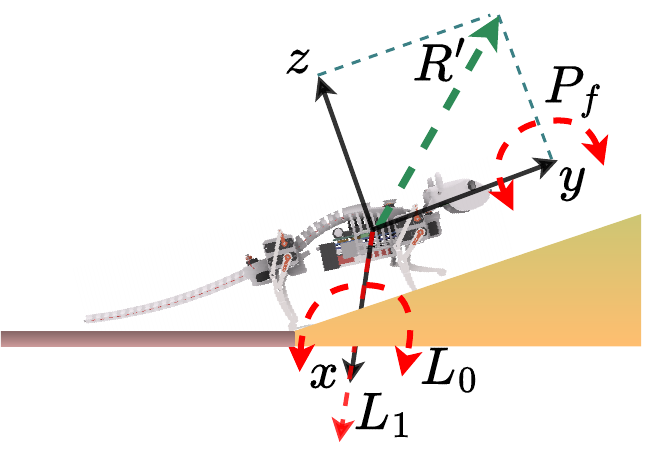}
		\label{fig:rampRE}
    }
    \subfigure[Reward function of spiral stairs environment.]{
		\includegraphics[width=.45\columnwidth]{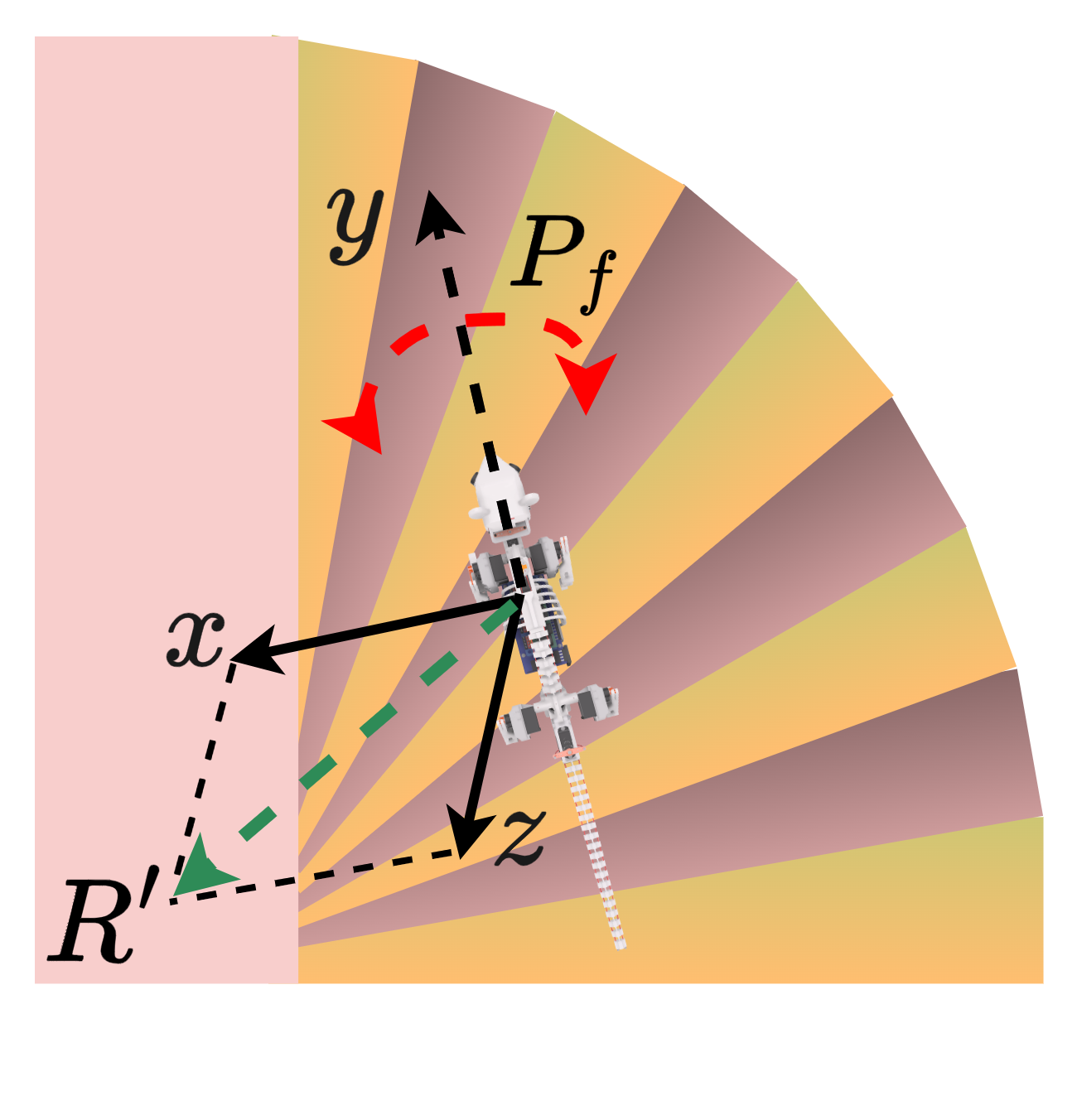}
		\label{fig:spiralRE}
    }
    \caption{Schematic of reward functions in different environments. The black arrows are NeRmo's local coordinates. The green arrows are the combined reward $R'$ of each scenario. The red arrows represent penalty terms and loss values. $P_f$ refers to the fall penalty. The loss values for unstable action or action with deviated direction are $L_0$ and $L_1$, respectively.}
    \label{fig:2Re}
\end{figure}
\begin{figure*}[ht]
    \centering
    \subfigure[Ramps.]{
        \includegraphics[width=.3\textwidth]{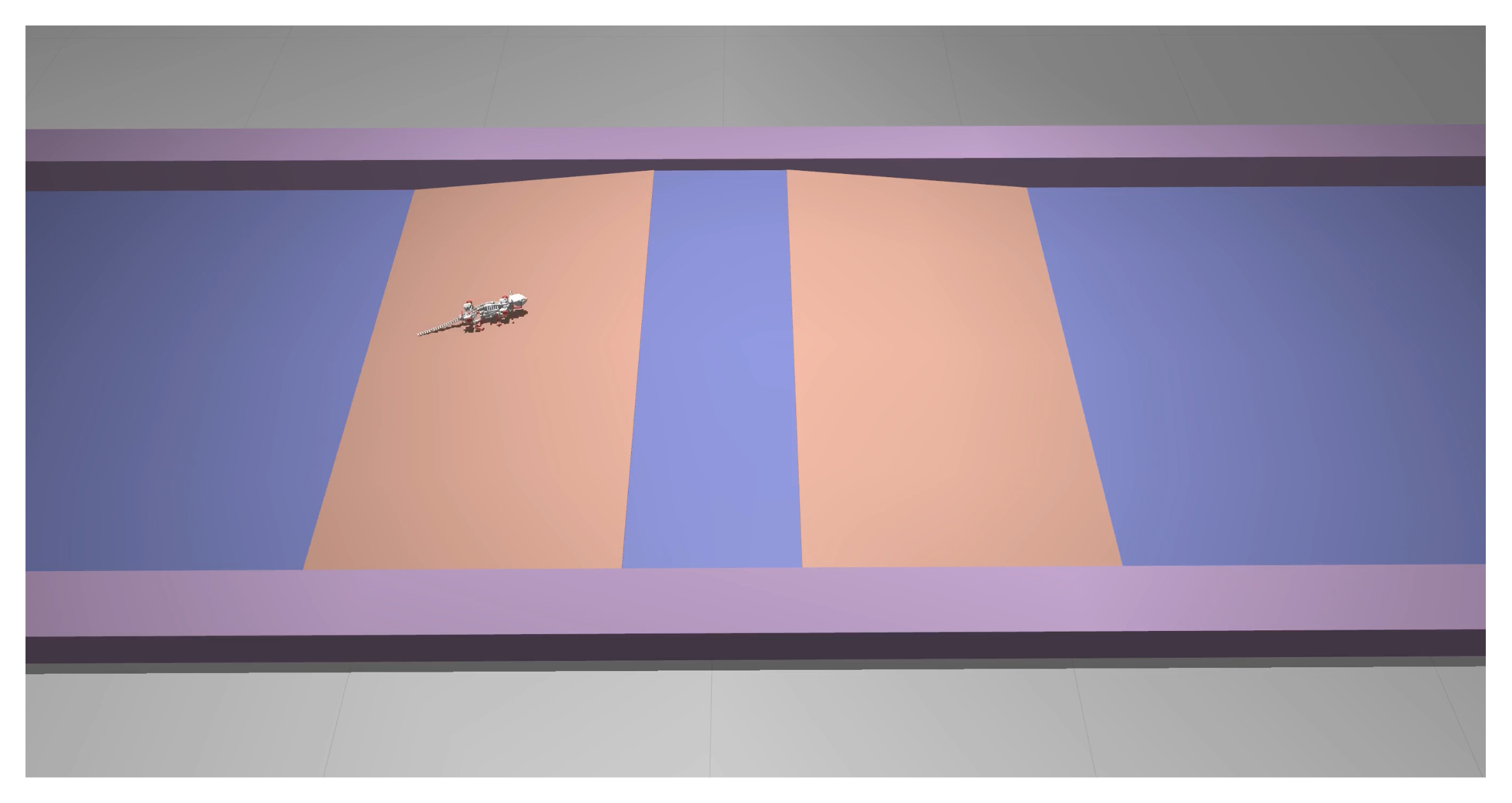}
        \label{fig:ramp_env}
    }
    \subfigure[Stairs.]{
	\includegraphics[width=.3\textwidth]{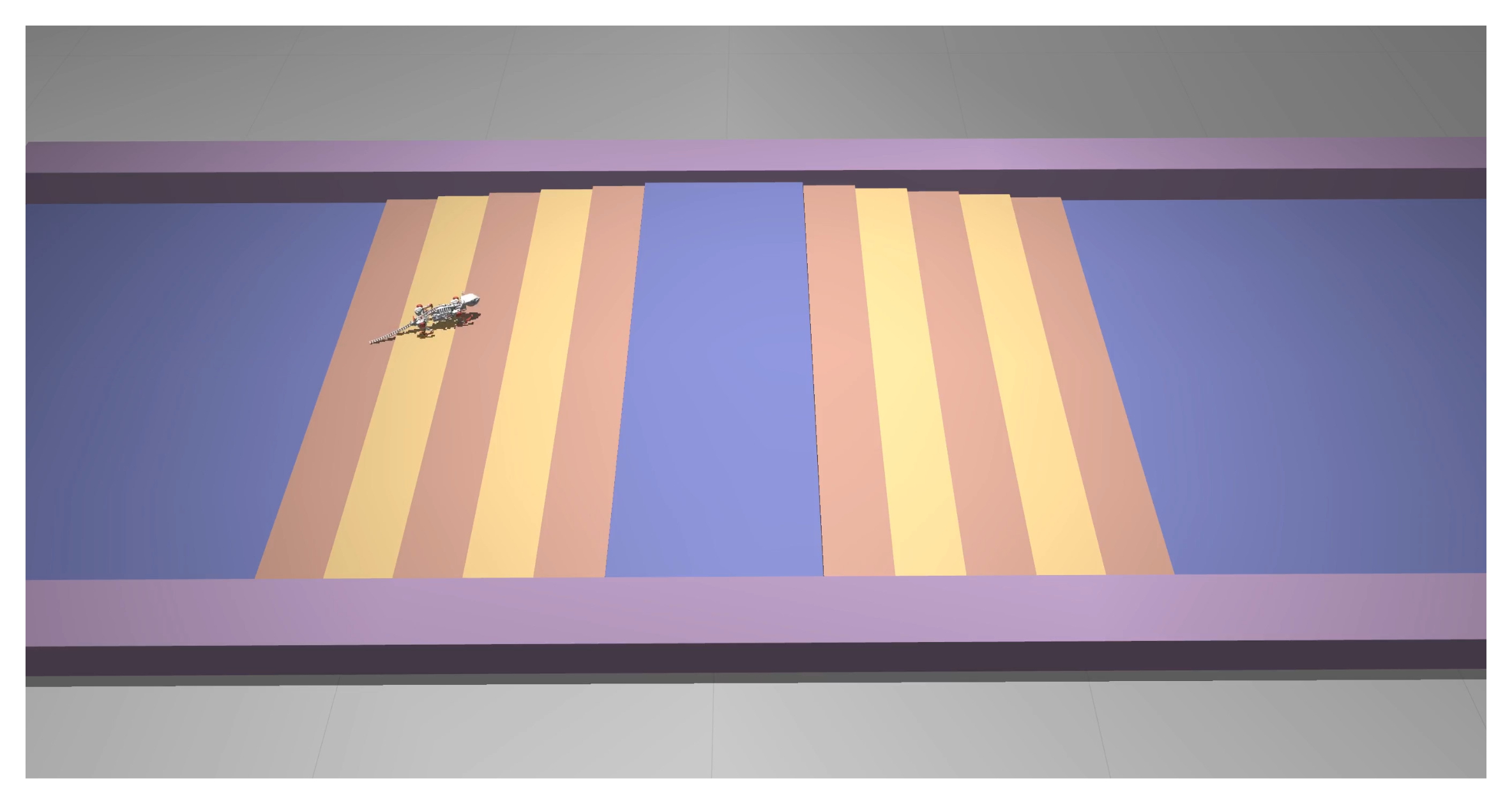}
        \label{fig:stairs_env}
    }
    \subfigure[Spiral Stairs.]{
	\includegraphics[width=.3\textwidth]{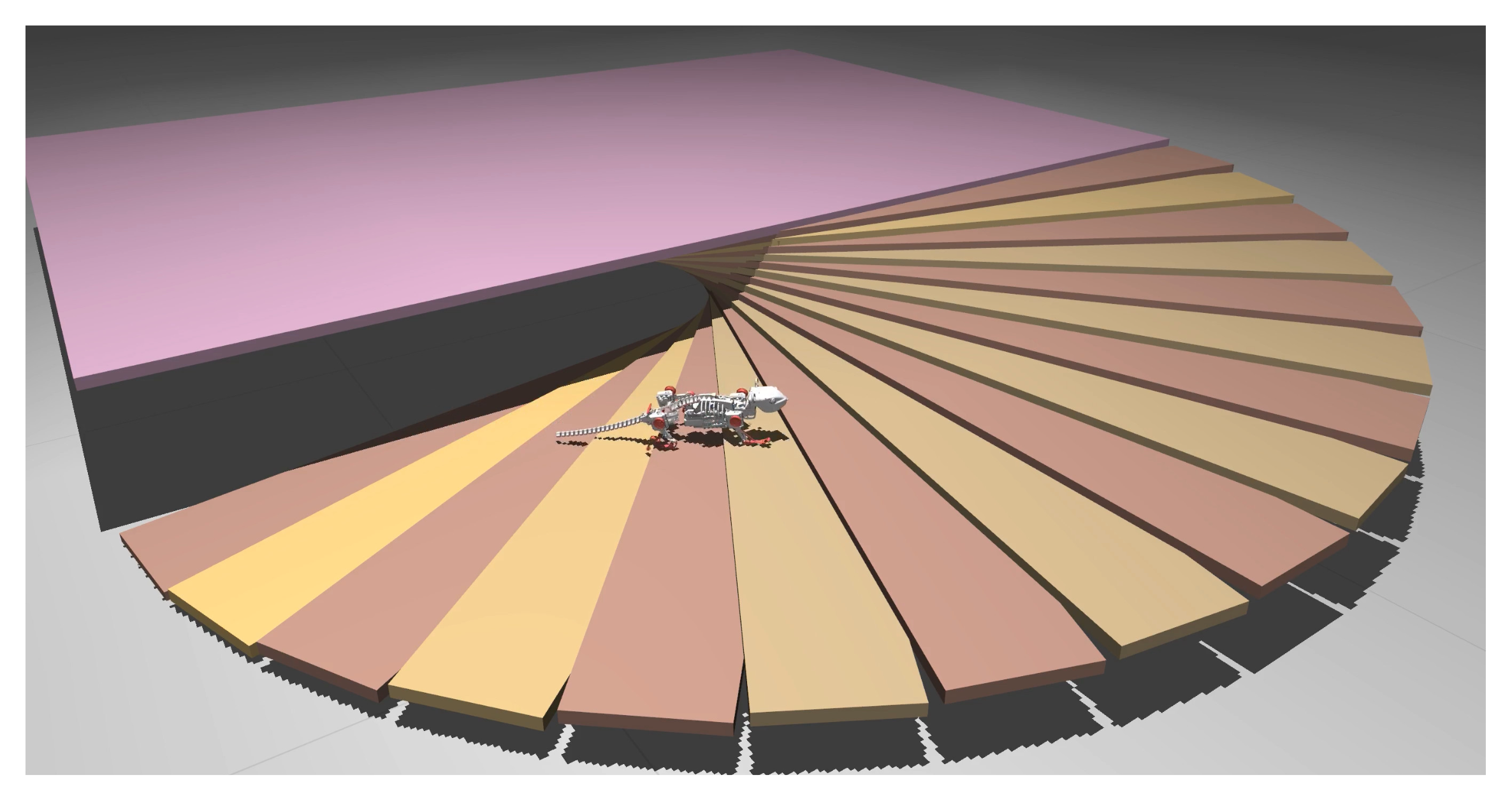}
        \label{fig:spiral_env}
    }
    \caption{Three kinds of hybrid environments. In (a) and (b), purple walls outline the walking zone of the rat robot.  (a)  is composed of two pink ramps and flat blue terrain. (b) and (c) are stair terrains, where steps are distinguished by different colors.} 
    \vspace{-0.5em}
    \label{fig:3env}
\end{figure*}

When the robot manages to walk continuously, the reward $R$ comprises both the loss values $\sum L_i$ and sub-rewards $(r(s_x), r(s_y), r(s_z))$ corresponding to each directional movement. To address unstable or off-target locomotion, several loss functions may be employed to indicate the necessity for correcting actions. Additionally, given the robot's works in a three-dimensional space, the combination of sub-rewards across each axis presents the objectives of any locomotion. Here, $s_x$, $s_y$, and $s_z$ denote the required sensory information associated with the sub-reward on the x-axis, y-axis, and z-axis, respectively. According to this, the sub-reward along a single axis is computed as:
\begin{equation}\label{reward_ge}
r(s) = 
\begin{cases} 
k \left( \frac{1}{1 + e^{-\alpha (s + \beta)}} - \gamma \right), &if \ required \\
0,& else
\end{cases}
\end{equation}
In \RefEq{reward_ge}, the parameter $k$ modulates the intensity of the reward, while $\alpha$ influences the optimization process. $\beta$ and $\gamma$ denote the horizontal and vertical offsets, respectively. The utilization of \RefEq{reward_ge} with customized parameters ensures that the reward gradient stabilizes upon reaching a predetermined threshold. This helps mitigate the undue influence of outlier actions and promotes stable learning of gait patterns. For instance, during stair ascent, rewards prioritize upward movement. On flat surfaces, forward progression is emphasized. This adaptive reward function enables the robot to autonomously navigate diverse tasks and environments, thereby enhancing its learning autonomy and adaptability.
As shown in \RefFig{fig:2Re}, two different scenarios considered in this paper are made as examples to explain the application of this reward function.

In the case of spiral stair scenarios, the robot is required to ascend the stairs while executing continuous turns. In this instance, the robot's angular velocity during turning and the upward displacement become the pertinent sensory information $(s_x, s_z)$. Here, $r(s_x)$ and $r(s_z)$ represent the rewards attributed to meeting the task target, and they are composed to construct $R'$ as illustrated in \RefFig{fig:spiralRE}.

In scenarios involving stairs and slopes, the robot's objective is to proceed forward and ascend. Hence, the robot's velocity in the forward direction or vertical movement serves as pertinent sensory information $(s_x, s_y)$ for reward computation. Here, $r(s_x)$ and $r(s_y)$ represent the respective rewards tied to achieving the required velocities, and they are combined to form $R'$ as illustrated in \RefFig{fig:rampRE}. Additionally, the robot is tasked with moving straight while minimizing lateral movements, penalizing deviations from the intended trajectory. In this context, the loss functions in \RefEq{re_all} are defined based on the robot's displacement along the x-axis and unstable states. Thus,
\begin{equation}\label{penalty}
\begin{aligned}
L_0 &= 
\begin{cases} 
-1,& \text{if } \theta_d > \theta_u \\
0,& else
\end{cases} \\
L_1& = -k'{((|l_x| + \delta)^{\alpha'} ) + \beta'}
\end{aligned}
\end{equation}
The angle $\theta_d$ denotes the deviation between the robot's movement directions at two continuous moments. If $\theta_d$ exceeds the predetermined threshold $\theta_u$, it signifies a significant shift in the robot's direction. Consequently, the robot's action is unstable, resulting in a substantial loss value assignment. The loss value $L_1$ presents the cumulative deviation of the robot's actions. Its primary objective is to shorten unnecessary deviations in the robot's movement, with larger deviations referring to higher penalties. In \RefEq{penalty}, $l_x$ represents the robot's displacement along the x-axis, with $\delta$ serving as a threshold to prevent excessive penalties for minor deviations. Parameters $k'$, $\alpha'$, and $\beta'$ dictate the trend of the loss value $L_1$.

By accounting for the robot's behaviors and enabling parameter-based adjustments, the proposed reward functions ensure relevance and flexibility across various tasks and environments, thereby enhancing the robot's capability to tackle diverse challenges.

\section{Experiment}\label{exp}
This section presents the simulation experiments designed to showcase the adaptability of the rat robot controlled by the proposed approach in complex environments. Specifically, three different types of environments will be employed to assess the effectiveness of the proposed RL approach in generating robot locomotion in diverse environments.

\subsection{Experimental Setup}
The small-size quadruped robot utilized in the experiment is the rat robot ``NeRmo'' as developed in our previous work \cite{bing2023lateral}. This robot measures merely 82mm in height, 96mm in width, and 375mm in length. We apply the Proximal Policy Optimization (PPO) Algorithms from stable baseline3 as the policy net in \RefFig{fig:archi}.
To evaluate the effectiveness of the proposed RL approach in generating adaptable locomotion across diverse environments, the rat robot is tasked with navigating through three distinct environments based on the learned locomotion. The specific tasks assigned to the robot within these environments (see \RefFig{fig:3env})are outlined in the following.
\begin{itemize}
    \item Ramps: In this scenario, the rat robot is required to ascend or descend ramps while walking straight. The slope of the ramps for both ascending and descending remains consistent. Across various experiments, the slope will be varied within the range of 5 degrees to 10 degrees.
    \item Stairs: In this scenario, the rat robot is tasked with ascending or descending stairs while maintaining a straight trajectory. The height of each step in various experiments ranges from 5 mm to 15 mm. Notably, 15 mm represents the maximum height of a gait stride achievable when the ``NeRmo'' is controlled using a model-based controller.
    \item Spiral Stairs: In this scenario, the rat robot is instructed to continuously change its direction while ascending the stairs.
\end{itemize}

\subsection{Training Process}

\begin{figure}[t]
    \centering
    \subfigure[Ramps.]{
	\includegraphics[width=.95\linewidth, trim=2.1cm 0.9cm 5cm 0.8cm, clip]{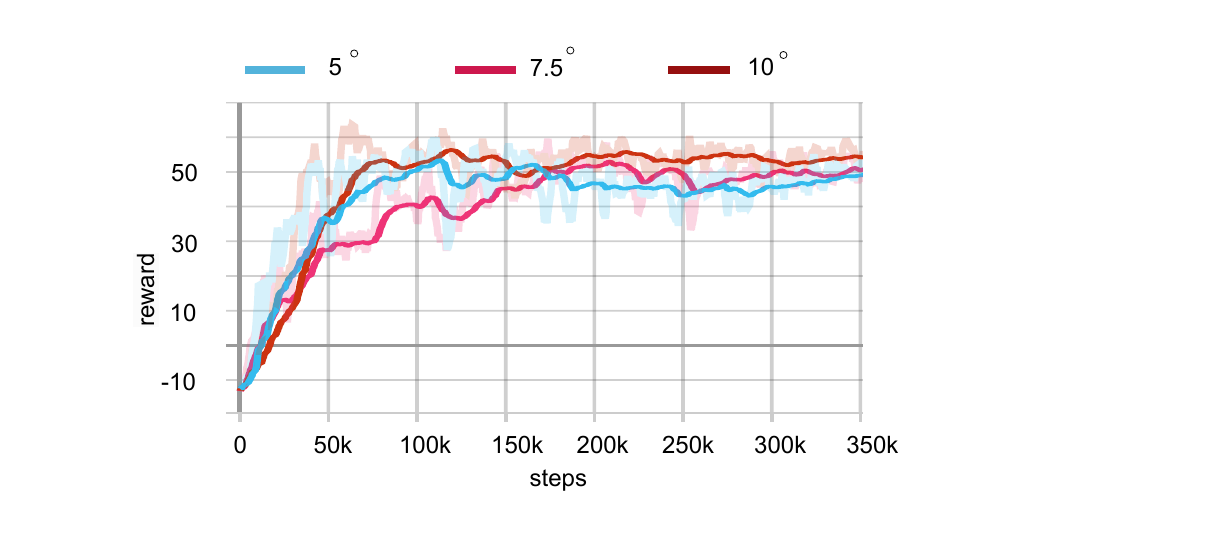}
        \label{fig:rampsR}
    }
    \subfigure[Stairs.]{
	\includegraphics[width=.95\linewidth, trim=2.1cm 0.9cm 5cm 0.2cm, clip]{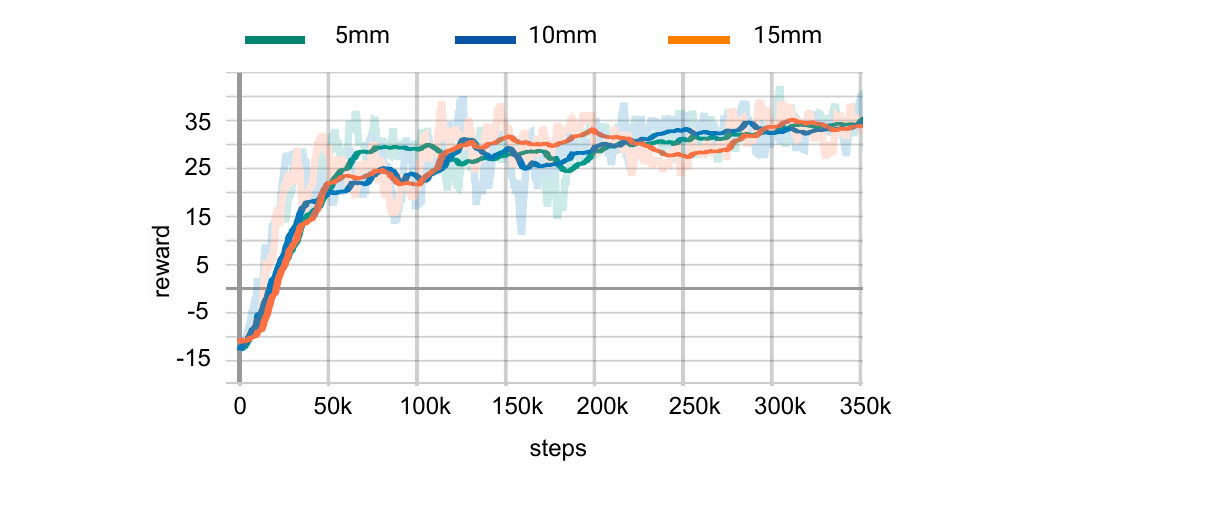} 
        \label{fig:stairsR}
    }
    \caption{The reward values of the proposed approach during training in different environments. (a) and (b) refers to the training results of \RefFig{fig:ramp_env} and \RefFig{fig:stairs_env} respectively. Lines with different colors in (a) refer to different slopes of the ramps. Lines with different colors in (b) involve the height of each step of the stairs.}
    \label{fig:re}
    \vspace{-0.5em}
\end{figure}

The training curves depicted in \RefFig{fig:re} demonstrate the convergence of action rewards across all tested scenarios. This convergence signifies that the proposed approach consistently deduces adaptable locomotion to effectively control the robot in response to diverse environmental changes. An interesting observation is that the rewards for robot actions in similar scenarios (as \RefFig{fig:rampsR} or \RefFig{fig:stairsR}) tend to converge to similar values. In such scenarios, the robots are tasked with similar objectives within analogous environmental conditions. The multifunctional reward mechanism devised in this paper accurately captures the interaction between the robot and its environment, facilitating the identification of nearly optimal control mechanisms to adapt to the environment. Consequently, rewards in similar scenarios exhibit similarity.

Additionally, \RefFig{fig:re} shows that training converges within 0.25M timesteps on both ramps or stairs. It must be acknowledged that directly comparing the learning processes of quadrupedal locomotion of different robots is quite challenging due to the differences in mechanical structures. We roughly compare the learning process of our approach with state-of-the-art approaches. We find that most approaches (such as \cite{bellegarda2022robust, luo2020carl, shi2022reinforcement}) endow robots with outstanding environmental adaptability. But they typically require iterations of at least several million time steps to converge and even require extensive pre-training efforts. In contrast, the proposed approaches can iteratively generate effective locomotion within 0.25M steps without any pre-training efforts. It is worth noting that, compared to state-of-the-art methods, the perceptual information of our robot is insufficient, which hinders RL from rapidly finding the optimal solution for environmental changes. However, by effectively extracting perceptual information, our approach enables RL to quickly identify various environmental changes, reducing the complexity of searching for the optimal solution and thereby improving its convergence rate. 

\subsection{Environmental Adaptability}
\begin{figure}[t]
    \centering
    \subfigure[Success Rate.]{
	\includegraphics[width=.85\linewidth]{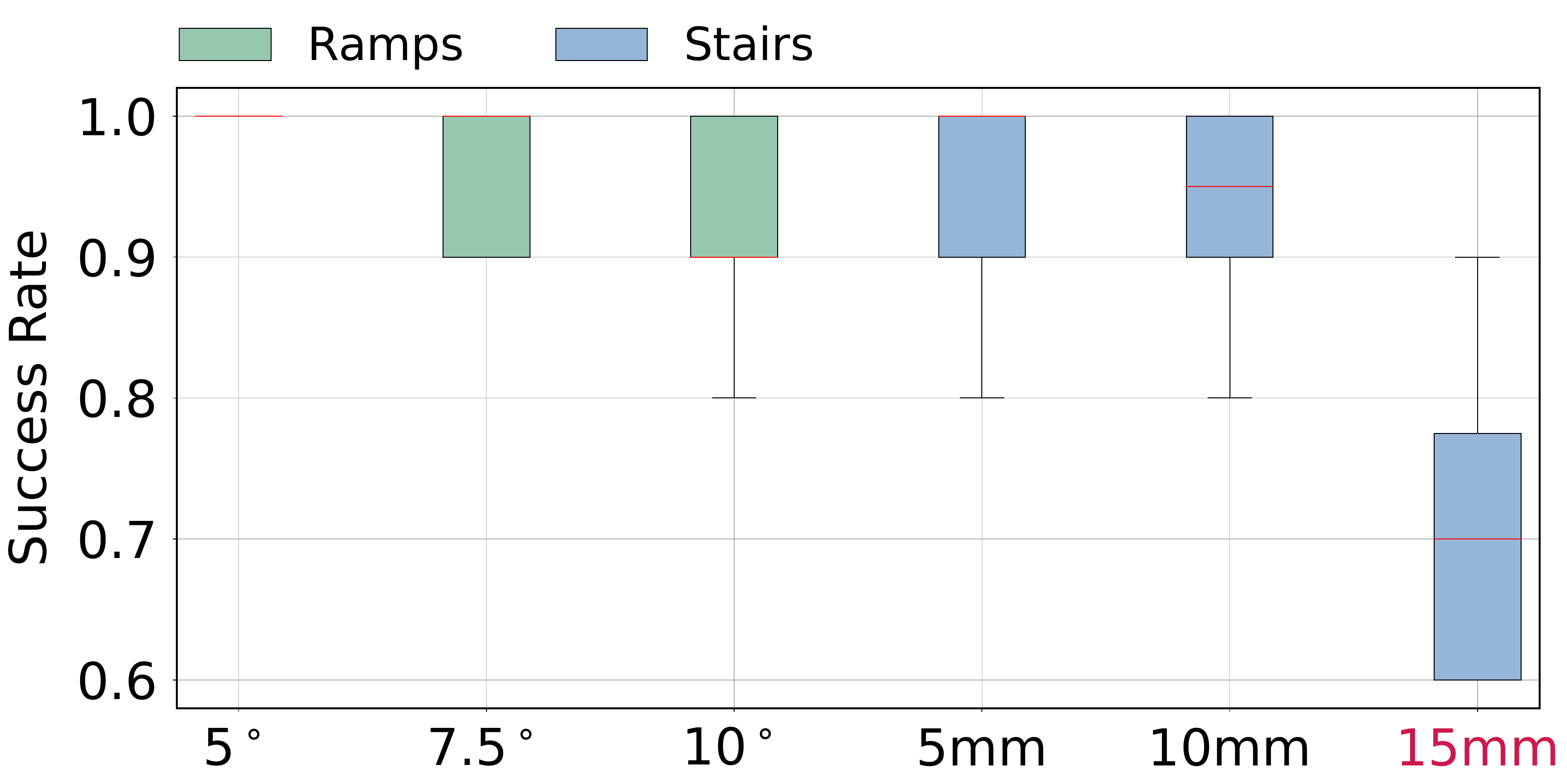}
        \label{fig:suc}
    }
    \subfigure[Time cost.]{
	\includegraphics[width=.85\linewidth]{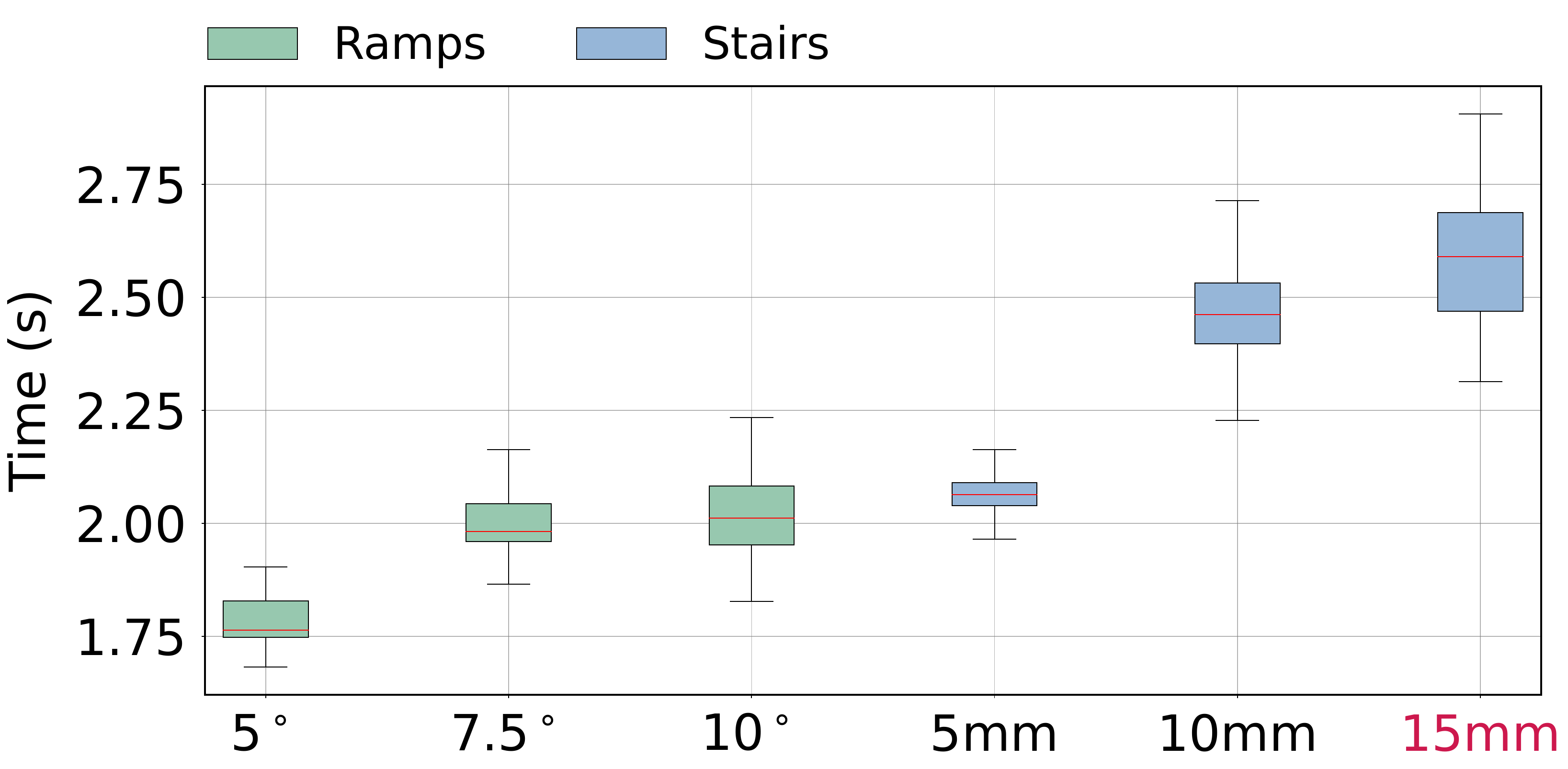}
        \label{fig:time}
    }
    \caption{Success rate and Time cost of tasks in diverse environments. Repeated experiments are conducted with the rat robot starting from random positions. Experimental errors are depicted in the boxes in the figure, with median values indicated by red lines.} 
    \label{fig:box}
    \vspace{-0.5em}
\end{figure}

In the experiments illustrated in \RefFig{fig:box}, the initial position of NeRmo is randomly selected within a predefined rectangular starting zone measuring 0.5m x 0.2m. The robot is instructed to advance a minimum distance of three meters within 60 gait strides. If the robot falls, exceeds the strides limit, or deviates significantly from the intended direction, the task is considered a failure. According to this, the success rate of task execution in diverse environments is defined. Additionally, for all successful cases, the time taken to complete the task is calculated to assess the performance of the learned locomotion.

\begin{figure*}[ht]
    \centering
    \subfigure[t = 0s,]{
	\includegraphics[width=.23\linewidth, trim=2cm 2cm 2cm 2cm, clip]{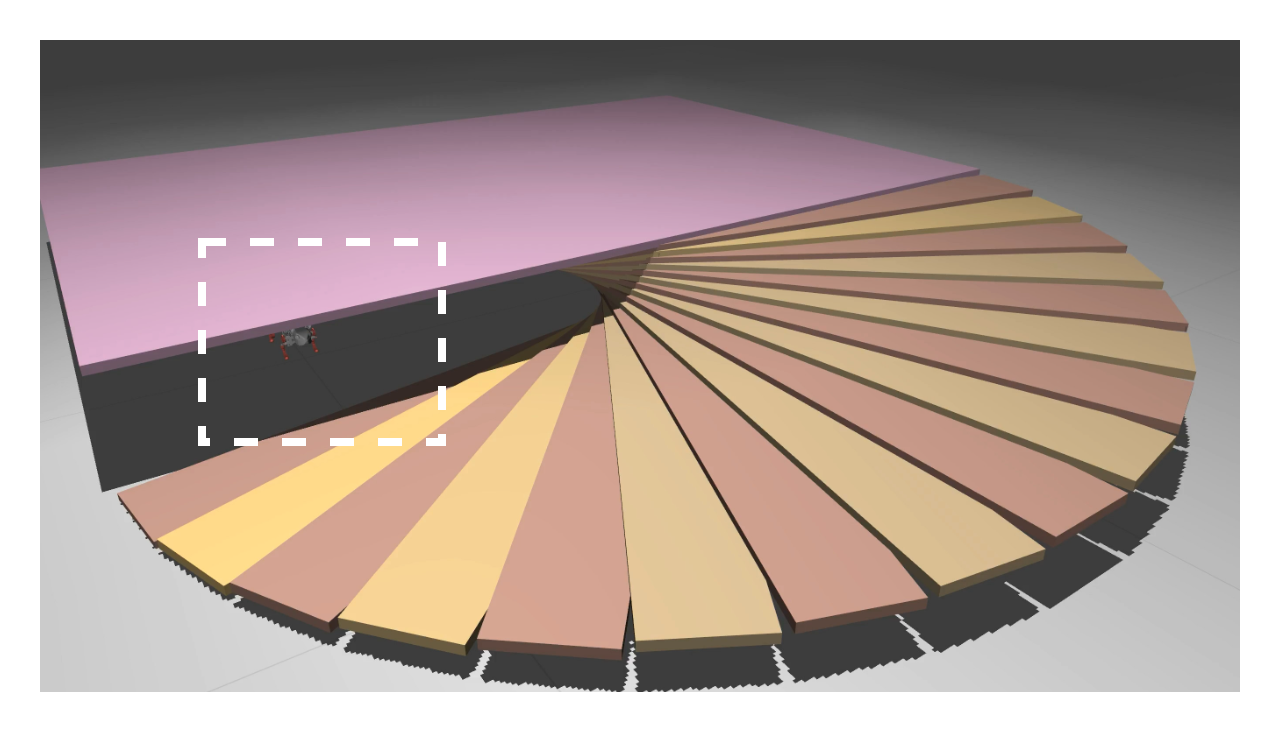}
        \label{fig:1sp}
    }\vspace{-0.1em} 
    \subfigure[t = 3s,]{
	\includegraphics[width=.23\linewidth, trim=2cm 2cm 2cm 2cm, clip]{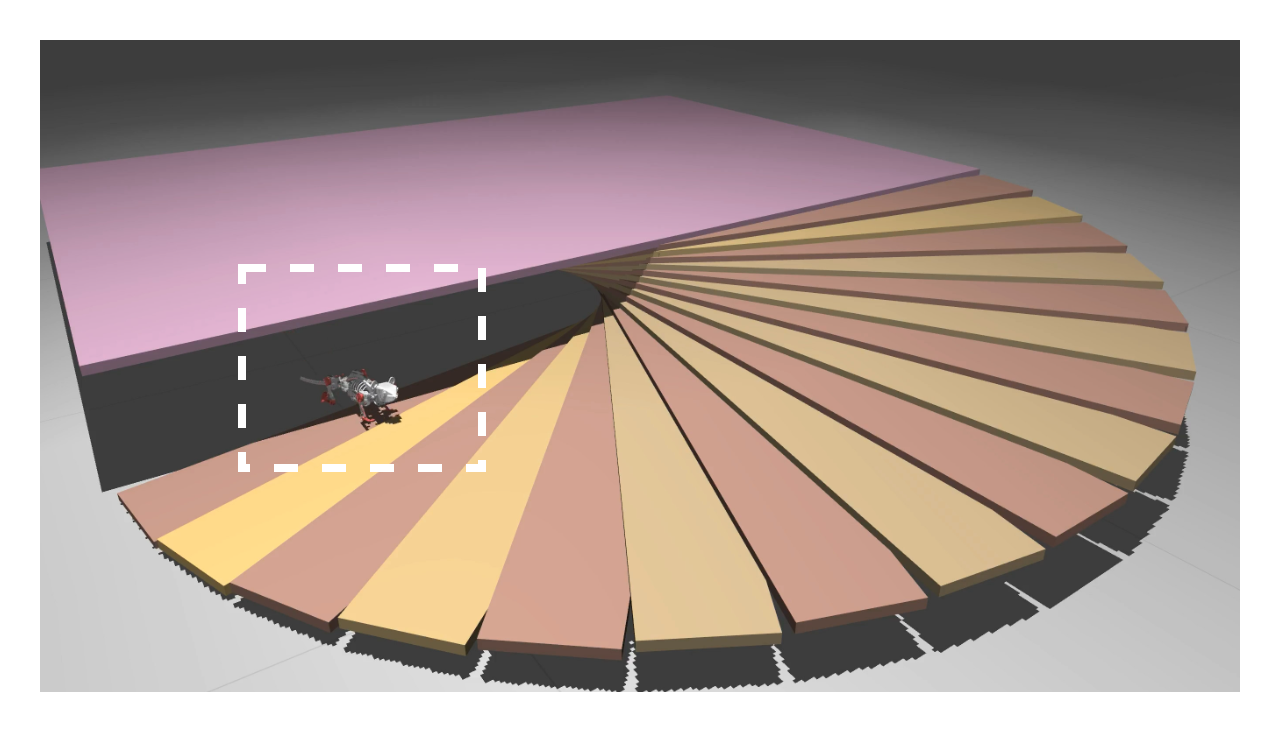}
        \label{fig:2sp}
    }\vspace{-0.1em} 
    \subfigure[t = 6s,]{
	\includegraphics[width=.23\linewidth, trim=2cm 2cm 2cm 2cm, clip]{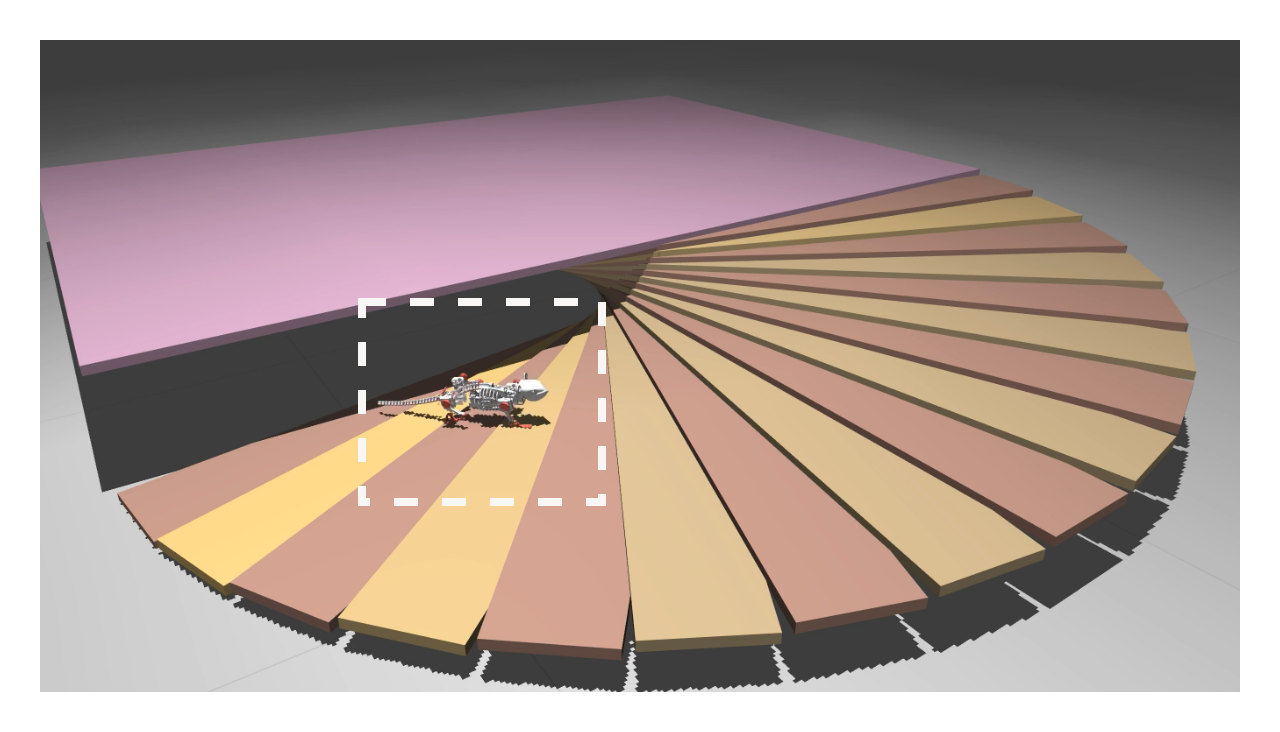}
        \label{fig:3sp}
    }\vspace{-0.1em} 
    \subfigure[t = 9s,]{
	\includegraphics[width=.23\linewidth, trim=2cm 2cm 2cm 2cm, clip]{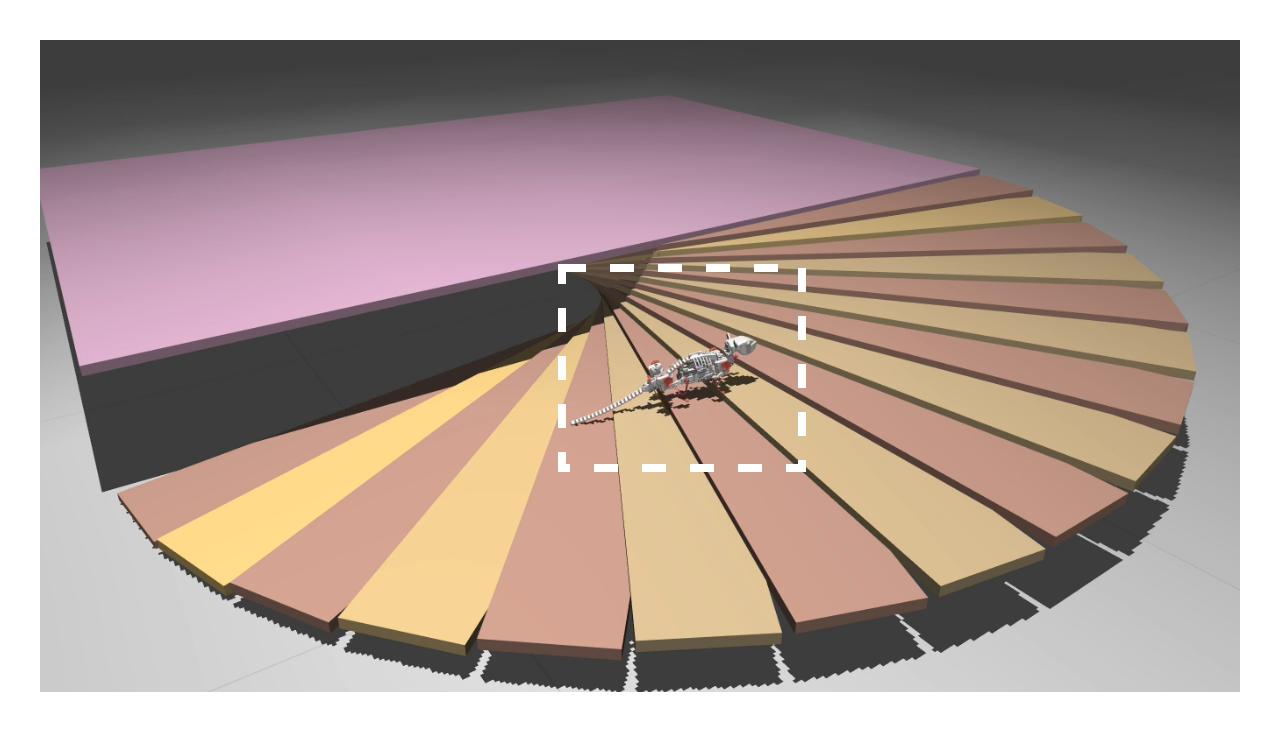}
        \label{fig:4sp}
    }\vspace{-0.1em} 
    \subfigure[t = 12s,]{
	\includegraphics[width=.23\linewidth, trim=2cm 2cm 2cm 2cm, clip]{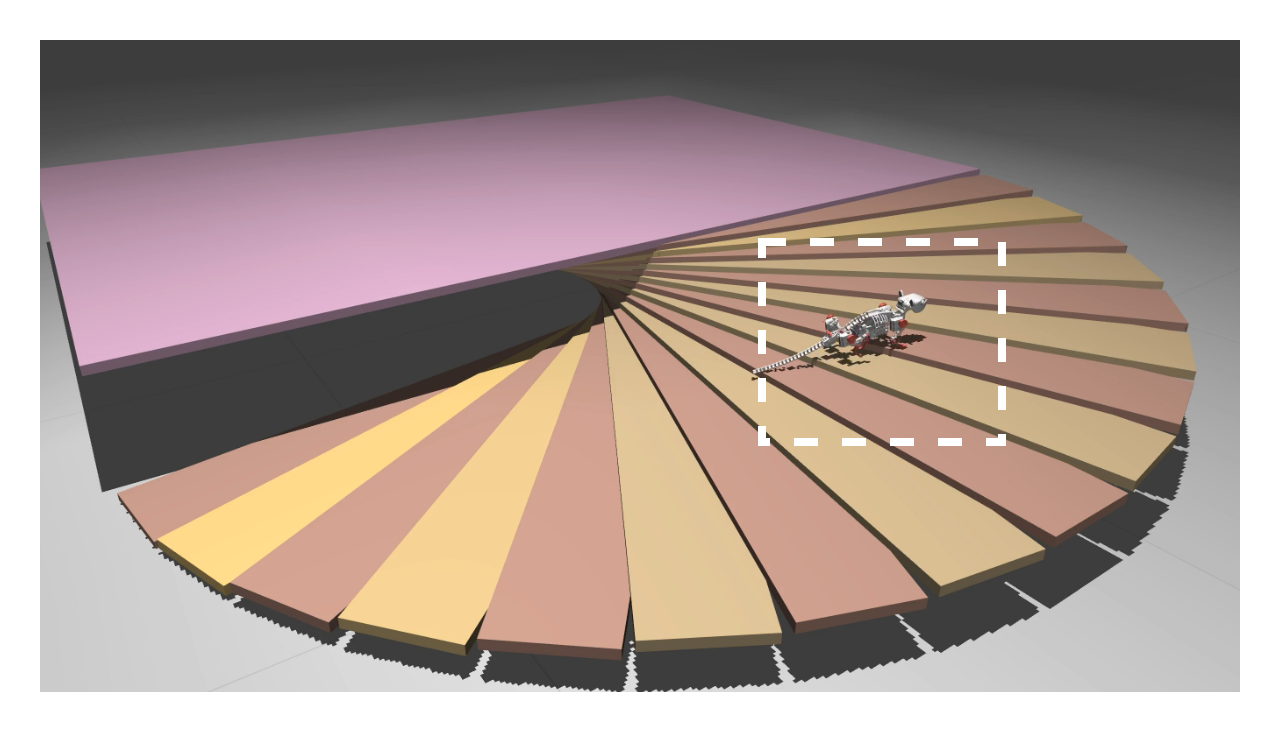}
        \label{fig:5sp}
    }\vspace{-0.1em} 
    \subfigure[t = 15s,]{
	\includegraphics[width=.23\linewidth, trim=2cm 2cm 2cm 2cm, clip]{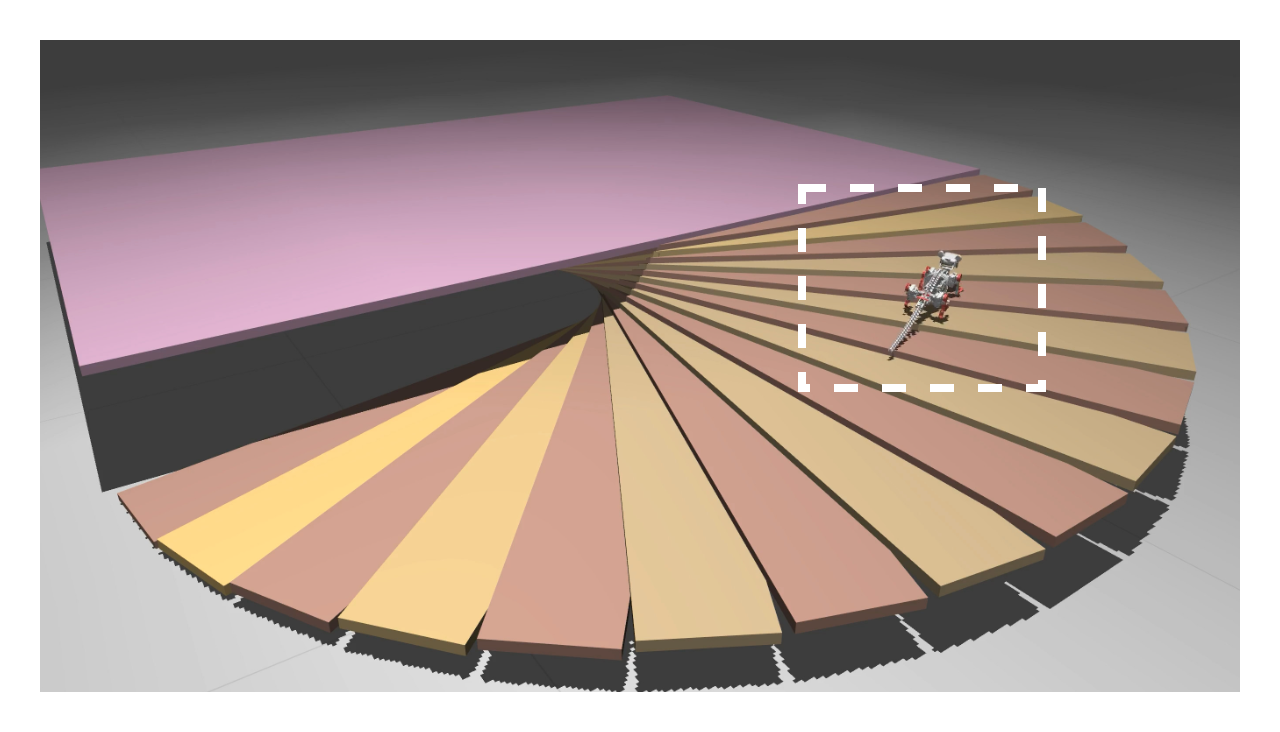}
        \label{fig:6sp}
    }\vspace{-0.1em} 
    \subfigure[t = 18s,]{
	\includegraphics[width=.23\linewidth, trim=2cm 2cm 2cm 2cm, clip]{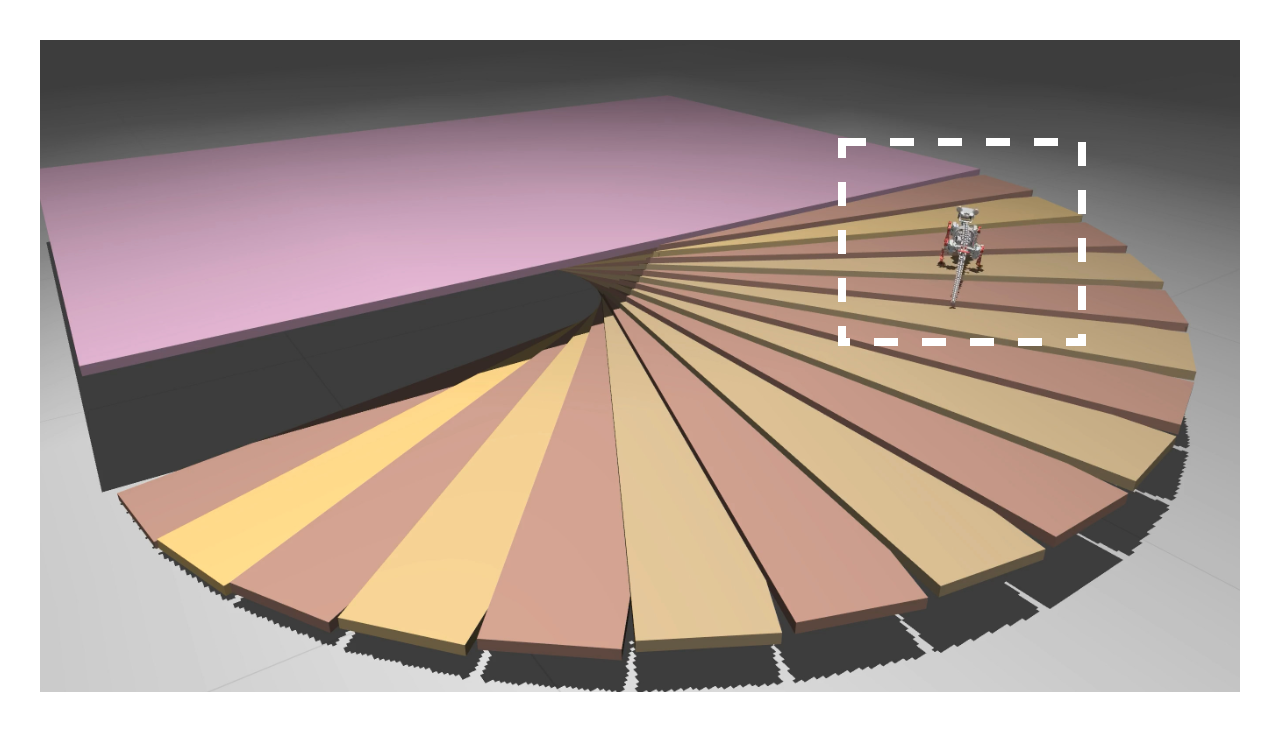}
        \label{fig:7sp}
    }\vspace{-0.1em} 
    \subfigure[t = 21s,]{
	\includegraphics[width=.23\linewidth, trim=2cm 2cm 2cm 2cm, clip]{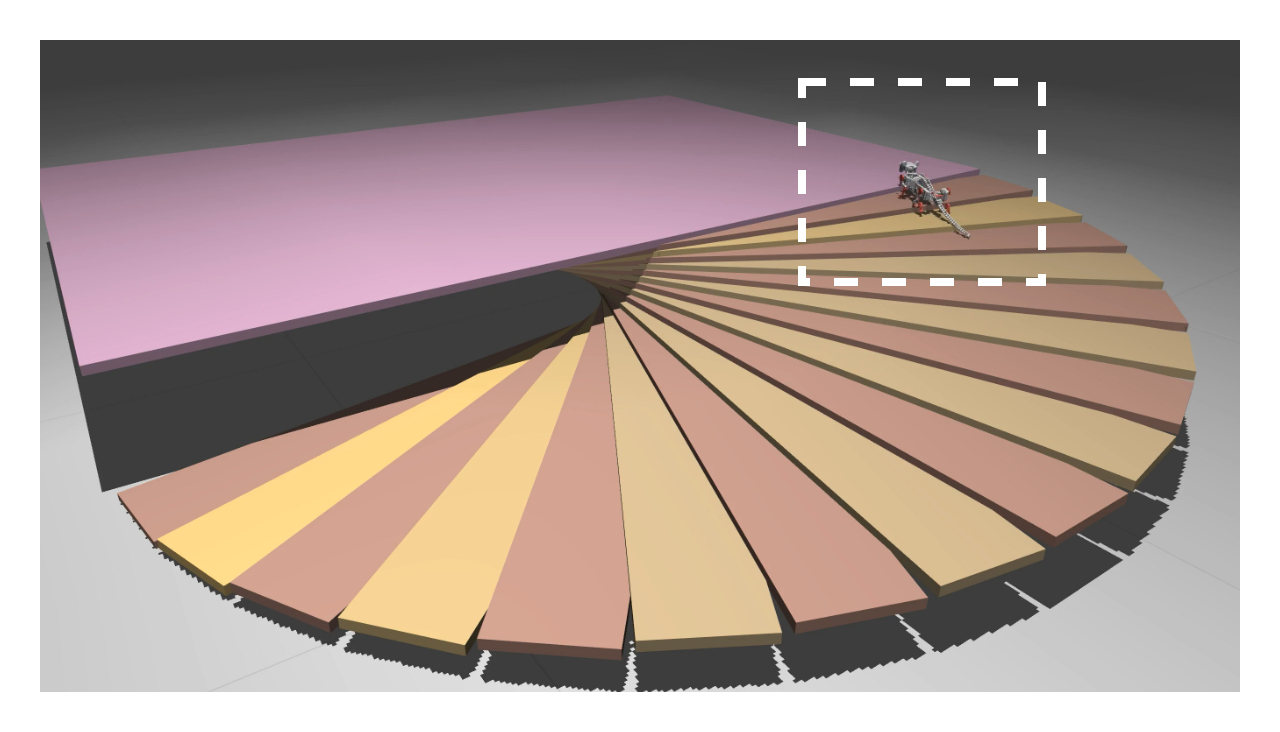}
        \label{fig:8sp}
    }\vspace{-0.1em} 
    \caption{Montage of NeRmo ascending spiral stairs. The white dashed rectangle shows NeRmo's real-time position on the spiral staircase.}
    \label{fig:spiral6}
    \vspace{-0.5em} 
\end{figure*}

\begin{figure}[t]
    \includegraphics[width=.95\linewidth, trim=2.1cm 0.8cm 5cm 0.2cm, clip]{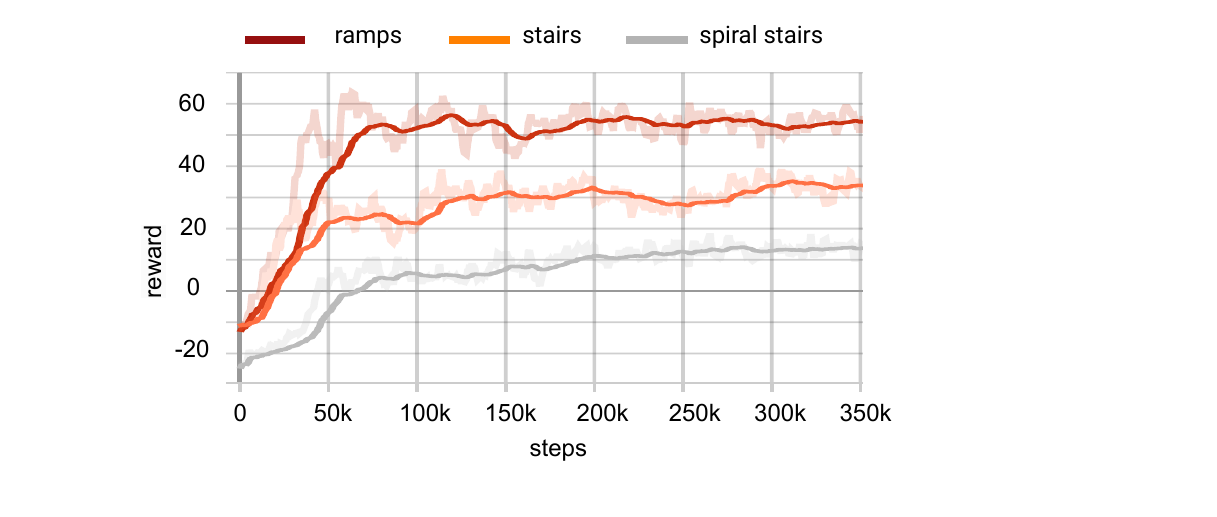}
    \vspace{-0.5em} 
    \caption{Comparison of rewards in different environments. Lines with different colors refer to the results of different scenarios.}
    \label{fig:3env_res}
    \vspace{-0.5em}
\end{figure}

\RefFig{fig:3env_res} illustrates that the increased task difficulty decreases the success rate of robot tasks, leading to longer time costs. This is a common occurrence, as the increasing complexity of the environment poses challenges for the robot to understand its surroundings based on reinforcement learning, while also disrupting the robot's movement. Nevertheless, with the exception of the stairs scenario with 15mm height steps, the robot consistently maintains a high success rate. This is credited to the ongoing enhancement of the robot's cognition of a particular environment through trial and error during reinforcement learning. In this context, the proposed approach enables the generation of optimal adaptable locomotion in diverse environments, leading to high success rates. For the stairs scenario with 15mm height steps, the robot faces extreme motion situations. This is due to the fact that 15 mm is the maximum height of a normal gait stride of NeRmo, limiting the robot's ability to adjust its locomotion in response to environmental changes when its initial position is altered. Despite a noticeable decline in the success rate, the NeRmo still achieves considerable success when encountering such extreme environmental conditions.




\subsection{Scenarios Extension}

To further evaluate the effectiveness of the proposed approach in generating adaptable locomotion across diverse environments, a challenging scenario is constructed using spiral stairs with stair heights of 15mm. In this setting, the robot is required not only to ascend the stairs but also to continuously adjust its direction. The rewards obtained during the training process of this scenario are depicted in  \RefFig{fig:3env_res}. Comparatively, the rewards for the spiral stairs scenario are notably lower than those for the other scenarios presented in \RefFig{fig:3env}, reflecting the heightened complexity of this environment. In other words, the designed multifunctional reward effectively captures environmental changes, indicating its generalizability. Furthermore, in \RefFig{fig:spiral6}, the movement of NeRmo during time as it ascends the spiral stairs is depicted. The robot's motion remains stable throughout, successfully reaching the final platform. The successful completion of this challenging task underscores the effectiveness and environmental adaptability of the proposed RL approach.



\section{Conclusion}\label{conclusion}
This paper presents a novel reinforcement learning approach for generating locomotion of small-scale quadruped robots in diverse environments. Due to limited size, these robots are equipped with limited and weak sensors, posing challenges in accurately perceiving environmental changes. To address this issue, our approach leverages the continuity and periodicity of robot motion to extract feedback information using Fourier transformation. Additionally, we propose a multifunctional reward mechanism aimed at learning locomotion capable of adapting to diverse environmental changes. The experimental results consistently demonstrate the effectiveness of our RL approach, with the robot maintaining stable mobility across different scenarios using only simple IMU data.


\bibliographystyle{IEEEtran}
\bibliography{iros24}

\begin{thebibliography}{10}
\providecommand{\url}[1]{#1}
\csname url@rmstyle\endcsname
\providecommand{\newblock}{\relax}
\providecommand{\bibinfo}[2]{#2}
\providecommand\BIBentrySTDinterwordspacing{\spaceskip=0pt\relax}
\providecommand\BIBentryALTinterwordstretchfactor{4}
\providecommand\BIBentryALTinterwordspacing{\spaceskip=\fontdimen2\font plus
\BIBentryALTinterwordstretchfactor\fontdimen3\font minus \fontdimen4\font\relax}
\providecommand\BIBforeignlanguage[2]{{%
\expandafter\ifx\csname l@#1\endcsname\relax
\typeout{** WARNING: IEEEtran.bst: No hyphenation pattern has been}%
\typeout{** loaded for the language `#1'. Using the pattern for}%
\typeout{** the default language instead.}%
\else
\language=\csname l@#1\endcsname
\fi
#2}}

\bibitem{gong2010review}
D.~Gong, J.~Yan, and G.~Zuo, ``A review of gait optimization based on evolutionary computation,'' \emph{Applied Computational Intelligence and Soft Computing}, vol. 2010, 2010.

\bibitem{biswal2021development}
P.~Biswal and P.~K. Mohanty, ``Development of quadruped walking robots: A review,'' \emph{Ain Shams Engineering Journal}, vol.~12, no.~2, pp. 2017--2031, 2021.

\bibitem{bellegarda2022robust}
G.~Bellegarda, Y.~Chen, Z.~Liu, and Q.~Nguyen, ``Robust high-speed running for quadruped robots via deep reinforcement learning,'' in \emph{2022 IEEE/RSJ International Conference on Intelligent Robots and Systems (IROS)}.\hskip 1em plus 0.5em minus 0.4em\relax IEEE, 2022, pp. 10\,364--10\,370.

\bibitem{degris2012model}
T.~Degris, P.~M. Pilarski, and R.~S. Sutton, ``Model-free reinforcement learning with continuous action in practice,'' in \emph{2012 American control conference (ACC)}.\hskip 1em plus 0.5em minus 0.4em\relax IEEE, 2012, pp. 2177--2182.

\bibitem{luo2020carl}
Y.-S. Luo, J.~H. Soeseno, T.~P.-C. Chen, and W.-C. Chen, ``Carl: Controllable agent with reinforcement learning for quadruped locomotion,'' \emph{ACM Transactions on Graphics (TOG)}, vol.~39, no.~4, pp. 38--1, 2020.

\bibitem{shi2022reinforcement}
H.~Shi, B.~Zhou, H.~Zeng, F.~Wang, Y.~Dong, J.~Li, K.~Wang, H.~Tian, and M.~Q.-H. Meng, ``Reinforcement learning with evolutionary trajectory generator: A general approach for quadrupedal locomotion,'' \emph{IEEE Robotics and Automation Letters}, vol.~7, no.~2, pp. 3085--3092, 2022.

\bibitem{aractingi2023controlling}
M.~Aractingi, P.-A. L{\'e}ziart, T.~Flayols, J.~Perez, T.~Silander, and P.~Sou{\`e}res, ``Controlling the solo12 quadruped robot with deep reinforcement learning,'' \emph{scientific Reports}, vol.~13, no.~1, p. 11945, 2023.

\bibitem{hwangbo2019learning}
J.~Hwangbo, J.~Lee, A.~Dosovitskiy, D.~Bellicoso, V.~Tsounis, V.~Koltun, and M.~Hutter, ``Learning agile and dynamic motor skills for legged robots,'' \emph{Science Robotics}, vol.~4, no.~26, p. eaau5872, 2019.

\bibitem{lee2020learning}
J.~Lee, J.~Hwangbo, L.~Wellhausen, V.~Koltun, and M.~Hutter, ``Learning quadrupedal locomotion over challenging terrain,'' \emph{Science robotics}, vol.~5, no.~47, p. eabc5986, 2020.

\bibitem{jain2019hierarchical}
D.~Jain, A.~Iscen, and K.~Caluwaerts, ``Hierarchical reinforcement learning for quadruped locomotion,'' in \emph{2019 IEEE/RSJ International Conference on Intelligent Robots and Systems (IROS)}.\hskip 1em plus 0.5em minus 0.4em\relax IEEE, 2019, pp. 7551--7557.

\bibitem{jenelten2024dtc}
F.~Jenelten, J.~He, F.~Farshidian, and M.~Hutter, ``Dtc: Deep tracking control,'' \emph{Science Robotics}, vol.~9, no.~86, p. eadh5401, 2024.

\bibitem{shi2020implementing}
Q.~Shi, Z.~Gao, G.~Jia, C.~Li, Q.~Huang, H.~Ishii, A.~Takanishi, and T.~Fukuda, ``Implementing rat-like motion for a small-sized biomimetic robot based on extraction of key movement joints,'' \emph{IEEE Transactions on Robotics}, vol.~37, no.~3, pp. 747--762, 2020.

\bibitem{bing2023lateral}
Z.~Bing, A.~Rohregger, F.~Walter, Y.~Huang, P.~Lucas, F.~O. Morin, K.~Huang, and A.~Knoll, ``Lateral flexion of a compliant spine improves motor performance in a bioinspired mouse robot,'' \emph{Science Robotics}, vol.~8, no.~85, p. eadg7165, 2023.

\bibitem{lucas2019development}
P.~Lucas, S.~Oota, J.~Conradt, and A.~Knoll, ``Development of the neurorobotic mouse,'' in \emph{2019 IEEE International Conference on Cyborg and Bionic Systems (CBS)}.\hskip 1em plus 0.5em minus 0.4em\relax IEEE, 2019, pp. 299--304.

\bibitem{huang2022enhanced}
Y.~Huang, Z.~Bing, F.~Walter, A.~Rohregger, Z.~Zhang, K.~Huang, F.~O. Morin, and A.~Knoll, ``Enhanced quadruped locomotion of a rat robot based on the lateral flexion of a soft actuated spine,'' in \emph{2022 IEEE/RSJ International Conference on Intelligent Robots and Systems (IROS)}.\hskip 1em plus 0.5em minus 0.4em\relax IEEE, 2022, pp. 2622--2627.

\bibitem{wang2024bioinspired}
R.~Wang, H.~Xiao, X.~Quan, J.~Gao, T.~Fukuda, and Q.~Shi, ``Bioinspired soft spine enables small-scale robotic rat to conquer challenging environments,'' \emph{Soft Robotics}, vol.~11, no.~1, pp. 70--84, 2024.

\end{thebibliography}

\end{document}